\documentclass{article}

\usepackage{microtype}
\usepackage{graphicx}
\usepackage{subfigure}
\usepackage{booktabs} 
\usepackage{amsmath}
\usepackage{amssymb}
\usepackage{enumitem}
\usepackage{verbatim}
\usepackage{placeins}
\usepackage{multicol}

\usepackage{hyperref}

\usepackage[accepted]{icml2020}

\icmltitlerunning{Interference and Generalization in Temporal Difference Learning}

\begin{document}

\twocolumn[
\icmltitle{Interference and Generalization in Temporal Difference Learning}



\icmlsetsymbol{equal}{*}

\begin{icmlauthorlist}
\icmlauthor{Emmanuel Bengio}{mcg,dmi}
\icmlauthor{Joelle Pineau}{mcg}
\icmlauthor{Doina Precup}{mcg,dm}
\end{icmlauthorlist}

\icmlaffiliation{mcg}{Mila, McGill University}
\icmlaffiliation{dm}{Deepmind}
\icmlaffiliation{dmi}{Work partly done while the author was an intern at Deepmind}

\icmlcorrespondingauthor{Emmanuel Bengio}{bengioe@gmail.com}

\icmlkeywords{Machine Learning, ICML, Reinforcement Learning}

\vskip 0.3in
]



\printAffiliationsAndNotice{}  

\begin{abstract}
We study the link between generalization and interference in temporal-difference (TD) learning. Interference is defined as the inner product of two different gradients, representing their alignment. This quantity emerges as being of interest from a variety of observations about neural networks, parameter sharing and the dynamics of learning. We find that TD easily leads to low-interference, under-generalizing parameters, while the effect seems reversed in supervised learning. We hypothesize that the cause can be traced back to the interplay between the dynamics of interference and bootstrapping. This is supported empirically by several observations: the negative relationship between the generalization gap and interference in TD, the negative effect of bootstrapping on interference and the local coherence of targets, and the contrast between the propagation rate of information in TD(0) versus TD($\lambda$) and regression tasks such as Monte-Carlo policy evaluation. We hope that these new findings can guide the future discovery of better bootstrapping methods.
\end{abstract}

\section{Introduction}
\label{intro}

The interference between two gradient-based processes, objectives $J_1,J_2$, sharing parameters $\theta$ is often characterized in the first order by the inner product of their gradients:
\begin{equation}
    \rho_{1,2} = \nabla_\theta J_1^T\nabla_\theta J_2\label{rho},
\end{equation}
and can be seen as the \emph{influence}, constructive ($\rho>0$) or destructive ($\rho<0$), of a gradient update using $\nabla_\theta J_1$ on $J_2$.

This quantity arises in a variety of ways (for completeness we rederive this quantity and others in appendix \ref{app:derivations}); it is the interference between tasks in multi-task and continual learning \citep{lopez2017gradient,schaul2019ray}, it forms the Neural Tangent Kernel \citep{jacot2018neural}, it is the Taylor expansion around $\theta$ of a gradient update \citep{achiam2019towards}, as well as the Taylor expansion of pointwise loss differences \citep{liu2019toward, fort2019stiffness}.

Interestingly, and as noted by works cited above, this quantity is intimately related to \emph{generalization}. If the interference between two processes is positive, then updating $\theta$ using gradients from one process will positively impact the other. Such processes can take many forms, for example, $J_1$ being the loss on training data and $J_2$ the loss on test data, or $J_1$ and $J_2$ being the loss on two i.i.d. samples.

The main claim of this work is that in Temporal Difference Learning (TD), interference evolves differently during training than in supervised learning (SL). More specifically, we find that \textbf{in TD learning lower interference correlates with a higher generalization gap while the opposite seems to be true in SL}, where low interference correlates with a low generalization gap (when early stopping).

In supervised learning, there is a wealth of literature suggesting that SGD has a regularization effect \citep[and references therein]{hardt2016train, zhang2016understanding, keskar2016large}, pushing the parameters in flat highly-connected \citep{draxler2018essentially, garipov2018LossSM} optimal regions of the loss landscape. It would be unsurprising for such regions of parameters to be on the threshold at the balance of bias and variance (in the traditional sense) and thus have low interference\footnote{\citet{fort2019stiffness} suggest that \emph{stiffness}, the cosine similarity of gradients, drops but stays positive once a model starts overfitting, it is also known that overfitting networks start having larger weights and thus larger gradients; this should result in the smallest $\rho$ precisely before overfitting happens.} as well as a low generalization gap.

Why is the situation then different with TD? This may be due to a multitude of factors. The evaluation methods of new algorithms in the recent union of neural networks and TD, despite an earlier recognition of the problem \citep{whiteson2011protecting}, often do not include generalization measures, perhaps leading to overfitting in algorithm space as well as solution space. This led to many works showing the brittleness of new TD methods \citep{machado2018revisiting, farebrother2018generalization,packer2018assessing}, and works proposing to train on a distribution of environments \citep{zhang2018study, cobbe2018quantifying, justesen2018illuminating} in order to have proper training and test sets \citep{zhang2018dissection,zhang2018natural}. 

In TD methods, models also face a different optimization procedure where different components may be at odds with each other, leading to phenomena like the deadly triad \citep{sutton2018reinforcement, achiam2019towards}. With most methods, from value-iteration to policy gradients, parameters are faced with an inherently non-stationary optimization landscape. In particular for value-based methods, bootstrapping induces an asymmetric flow of information (from newly explored states to known states) which remains largely unexplored in deep learning literature. Such non-stationarity and asymmetry may help explain the success of sparse methods \citep{sutton1996generalization, liu2019utility} that act more like tabular algorithms (with convergence at the cost of more updates). 

Other works also underline the importance of interference. \citet{riemer2018learning} show that by simply optimizing for interference accross tasks via a naive meta-learning approach, one can improve RL performance. Interestingly, \citet{nichol2018first} also show how popular meta-learning methods implicitly also maximize interference. Considering that the meta-learning problem is inherently interested in generalization, this also suggests that increasing constructive interference should be beneficial.

Why then do TD methods naturally induce under-generalizing low-interference solutions? We first offer an empirical investigation confirming this behavior. Then, we reinterpret results on popular environments where generalization is not typically measured, showing that models may very well be in memorization-mode and lack temporal coherence. Finally, we attempt to offer some mathematical insights into this phenomenon.

\section{Preliminaries}

A Markov Decision Process (MDP)~\citep{bellman1957markovian,sutton2018reinforcement} $\mathcal{M}=\left<S,A,R,P,\gamma\right>$ consists of a state space $S$, an action space $A$, a reward function $R:S\to\mathbb{R}$ and a transition probability distribution $P(s'|s,a)$. RL agents aim to optimize the long-term return,
\begin{align*}
    G(S_t) &= \sum_{k=t}^\infty\gamma^{k-t}R(S_{k}), \label{eq:mc-return}
\end{align*}
in expectation, where $\gamma\in [0,1)$ is called the discount factor.
Policies $\pi(a|s)$ map states to action distributions. Value functions $V^{\pi}$ and $Q^{\pi}$ map states/states-action pairs to expected returns, and can be expressed recursively:
\begingroup
\allowdisplaybreaks
\begin{align*}
    V^\pi(S_t) &= \mathbb{E}_{\pi}[G(S_t)] 
    \\&=\mathbb{E}_{\pi}[R(S_{t}) + \gamma V(S_{t+1}) |A_t \sim \pi(S_t)]\\
    Q^\pi(S_t, A_t) &= \mathbb{E}_{\pi}[R(S_{t}) + \gamma\sum_a \pi(a|S_{t+1}) Q(S_{t+1}, a)]
\end{align*}
\endgroup
While $V^\pi$ could also be learned via regression to observed values of $G$, these recursive equations give rise to the \emph{Temporal Difference (TD)}  update rules for policy evaluation, relying on current estimates of $V$ to \emph{bootstrap}, e.g.:
\begin{equation}
    V(S_t) \leftarrow V(S_t) - \alpha(V(S_t) - (R(S_t) + \gamma V(S_{t+1}))),
\end{equation}
where $\alpha\in[0,1)$ is the step-size.
Bootstrapping leads also to algorithms such as \textbf{Q-Learning}~\citep{watkins1992q}:
\begin{equation}
    \mathcal{L}_{QL} = [Q_\theta(S_t,A_t) - (R_t + \gamma \max_{a}Q_{\theta'}(S_{t+1},a))]^2,
    \label{eq:ql}
\end{equation}


fitted-Q~\citep{ernst2005tree,riedmiller2005neural},
and \textbf{TD($\mathbf{\lambda}$)}, which trades off between the unbiased target $G(S_t)$ and the biased  TD(0) target (biased due to  relying on the estimated $V(S_{t+1})$),  using a weighted averaging of future targets called a $\lambda$-return~\citep{sutton1988learning, munos2016safe}:
\begin{align}
G^\lambda(S_t) &=(1-\lambda)\sum\nolimits_{n=1}^{\infty} \lambda^{n-1}G^{n}(S_t)\\
G^{n}(S_t) &= \gamma^n V(S_{t+n}) + \sum\nolimits_{j=0}^{n-1} \gamma^jR(S_{t+j}) \nonumber\\
\mathcal{L}_{TD(\lambda)}(S_t) &= (V_\theta(S_t) - G^\lambda(S_t))^2, \label{eq:td_lambda}
\end{align}
(note that the return depends implicitly on the trajectory and the actions followed from $S_t$).
When $\lambda=0$, the loss is simply $(V_\theta(S_t)-(R_t+\gamma V_\theta(S_{t+1})))^2$, leading to the TD(0) algorithm \citep{sutton1988learning}.

Learning how to act can be done solely using a value function, e.g. with the greedy policy $\pi(s) = \mbox{argmax}_a Q(s,a)$. Alternatively, one can directly parameterize the policy as a conditional distribution over actions, $\pi_\theta(a|s)$. In this case, $\pi$ can be updated directly with policy gradient (PG) methods, the simplest of which is REINFORCE \citep{williams_simple_1992}:
\begin{align}
    \nabla_\theta G(S_t) = G(S_t) \nabla_\theta \log \pi(A_t| S_t). \label{eq:reinforce}
\end{align}

\subsection{Computing interference quantities}
\label{sec:comp-interf}

Comparing loss interference in the RL and SL case isn't necessarily indicative of the right trends, due to the fact that in most RL algorithms, the loss landscape itself evolves over time as the policy changes. Instead, we remark that loss interference, $\rho_{1,2} = \nabla_\theta J_1^T\nabla_\theta J_2$, can be decomposed as follows. Let $J$ be a scalar loss, $u$ and $v$ some examples: 
\begin{align}
    \rho_{u,v} = \frac{\partial J(u)}{\partial f(u)}\frac{\partial f(u)}{\partial \theta}^T\frac{\partial f(v)}{\partial \theta}\frac{\partial J(v)}{\partial f(v)}.
\end{align}
While the partial derivative of the loss w.r.t. $f$ may change as the loss changes, we find experimentally that the inner product of gradients of the output of $f$ remains stable\footnote{Although gradients do not always converge to 0, at convergence the parameters themselves tend to \emph{wiggle around} a minima, and as such do not affect the \emph{function} and its derivatives that much.}. As such, we will also compute this quantity throughout, function interference, as it is more stable and reflects interference at the representational level rather than directly in relation to the loss function:
\begin{align}
    \bar{\rho}_{u,v} = \frac{\partial f(u)}{\partial \theta}^T\frac{\partial f(v)}{\partial \theta}.
\end{align}
For functions with more than one output in this work, e.g. a softmax classifier, we consider the output, $f(u)$, to be the max\footnote{This avoids computing the (expensive) Jacobian, we also find that this simplification accurately reflects the same trends experimentally.}, e.g. the confidence of the argmax class.

\section{Empirical Setup}

We loosely follow the setup of \citet{zhang2018dissection} for our generalization experiments: we train RL agents in environments where the initial state is induced by a single random seed, allowing us to have proper training and test sets in the form of mutually exclusive seeds. In particular, to allow for closer comparisons between RL and SL, we compare classifiers trained on SVHN \citep{Netzer2011ReadingDI} and CIFAR10 \citep{Krizhevsky2009LearningML} to agents that learn to progressively explore a masked image (from those datasets) while attempting to classify it. The random seed in both cases is the index of the example in the training or test set.

More specifically, agents start by observing only the center, an $8\times 8$ window of the current image. At each timestep they can choose from 4 movement actions, moving the observation window by 8 pixels and revealing more of the image, as well as choose from 10 classification actions. The episode ends upon a correct classification or after 20 steps. 

We train both RL and SL models with the same architectures, and train RL agents with a Double DQN objective \citep{hasselt2015deep}. We also train REINFORCE \citep{williams_simple_1992} agents as a test to entirely remove dependence on value estimation and have a pure Policy Gradient (PG) method. 

As much of the existing deep learning literature on generalization focuses on classifiers, but estimating value functions is arguably closer to regression, we include two regression experiments using SARCOS \citep{vijayakumar2000locally} and the California Housing dataset \citep{KELLEYPACE1997291}.

Finally, we investigate some metrics on the popular Atari environment \citep{bellemare2013arcade} by training DQN \citep{mnih2013playing} agents, with the stochastic setup recommended by \citet{machado2018revisiting}, and by performing policy evaluation with the Q-Learning and TD($\lambda$) objectives. To generate expert trajectories for policy evaluation, we run an agent pretrained with Rainbow~\citep{hessel2018rainbow}; we denote $\mathcal{D}^*$ a dataset of transitions obtained with this agent, and $\theta^*$ the parameters after training that agent.

We measure correlations throughout with Pearson's $r$:
\begin{equation}
   r_{xy} =\frac{\sum ^n _{i=1}(x_i - \bar{x})(y_i - \bar{y})}{\sqrt{\sum ^n _{i=1}(x_i - \bar{x})^2} \sqrt{\sum ^n _{i=1}(y_i - \bar{y})^2}}
\end{equation}
which is a measure, between $-1$ and $1$, of the linear correlation between two random variables $X,Y$.
All architectural details and hyperparameter ranges are listed in appendix \ref{app:hyperparams}.

\section{Empirical observations of interference and generalization}
To measure interference in the overparameterized regime and still be able to run many experiments to obtain trends, we instead reduce the number of training samples while also varying capacity (number of hidden units and layers) with smaller-than-state-of-the-art but reasonable architectures. Two results emerge from this.

\begin{figure}[hb]
\vskip 0.2in
\begin{center}
\centerline{\includegraphics[width=\columnwidth]{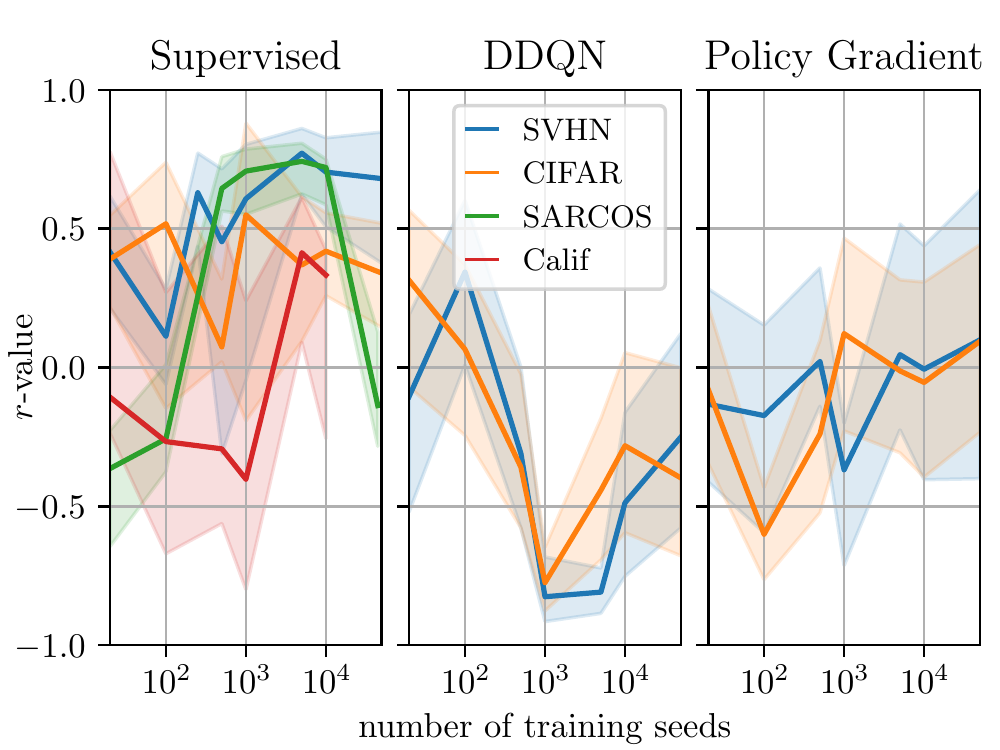}}
\caption{Correlation coefficient $r$ between the (log) function interference $\bar{\rho}$ and the generalization gap, as a function of training set size; shaded regions are bootstrapped 90\% confidence intervals. We see different trends for value-based experiments (middle) than for supervised (left) and PG experiments (right).}
\label{corr_vs_ntrain}
\end{center}
\vskip -0.2in
\end{figure}

First, in Fig. \ref{corr_vs_ntrain} for each training set size, we measure the correlation between interference and the generalization gap. We see that, after being given sufficient amounts of data, TD methods tend to have a strong negative correlation, while classification methods tend to have positive correlation. 

Regression has similar but less consistent results; SARCOS has a high correlation peak when there starts being enough data, albeit shows no correlation at all when all 44k training examples are given (the generalization gap is then almost 0 for all hyperparameters); on the other hand the California dataset only shows positive correlation when most or all of the dataset is given. The trends for PG SVHN and CIFAR show no strong correlations (we note that $|r|<0.3$ is normally considered to be a weak correlation; \citealp{cohen2013statistical}) except for PG CIFAR at 100 training seeds, with $r=-.60$.

\begin{figure}[ht]
\vskip 0.2in
\begin{center}
\centerline{\includegraphics[width=\columnwidth]{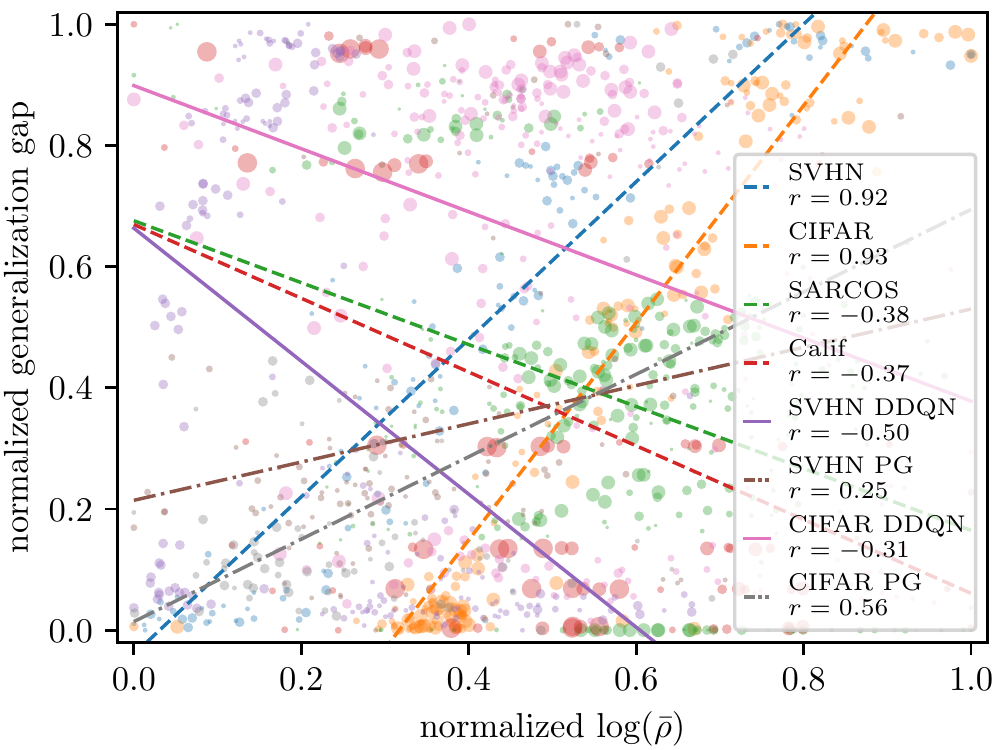}}
\caption{Generalization gap vs interference $\bar{\rho}$ for all runs. Larger circles represent larger capacity models. Here value-based methods seem to be behaving like regression methods.}
\label{norm_gen_gap_vs_inter}
\end{center}
\vskip -0.2in
\end{figure}

Second, in Fig. \ref{norm_gen_gap_vs_inter}, we plot the generalization gap against interference $\bar{\rho}$ for every experiment (normalized for comparison). We then draw the linear regression for each experiment over all training set sizes and capacities.  For both classification tasks, interference is strongly correlated ($r>0.9$) with the generalization gap, and also is to a lesser extent for the PG experiments. For all other experiments, regression and value-based, the correlation is instead negative, albeit low enough that a clear trend cannot be extracted. Note that the generalization gap itself is almost entirely driven by the training set size first ($r<-0.91$ for all experiments except PG, where $r$ is slightly higher, see appendix Fig. \ref{gen_gap_vs_ntrain}).

The combination of these results tells us that not only does interference evolves differently in TD than in SL, it has some similarities with regression, as well as a different characterization of memorization: \textbf{in classification, low-interference solutions tend to generalize, while in TD, low-interference solutions often memorize}. In regression, this seems only true for a fixed quantity of data.

\section{Interference in Atari domains}

The Arcade Learning Environment \citep{bellemare2013arcade}, comprised of Atari games, has been a standard Deep RL benchmark in the recent past \citep{mnih2013playing,bellemare2017distributional,kapturowski2018recurrent}. 
We once again revisit this benchmark to provide additional evidence of the memorization-like behaviors of value-based methods on these domains. Understanding the source of these behaviors is important, as presumably algorithms may be able to learn generalizing agents from the same data. Additionally, such low-interference memorization behaviors are not conducive to sample efficiency, which even in an environment like Atari, could be improved.

Recall that interference is a first order Taylor expansion of the pointwise loss difference, $J_{\theta'} - J_\theta$. Evaluating such a loss difference is more convenient to do on a large scale and for many runs, as it does not require computing individual gradients. In this section, we evaluate the expected TD loss difference for several different training objectives, a set of supervised objectives, the Q-Learning objective applied first as policy evaluation (learning from a replay buffer of expert trajectories) and then as a control (learning to play from scratch) objective, and the TD($\lambda$) objective applied on policy evaluation. Experiments are ran on MsPacman, Asterix, and Seaquest for 10 runs each. Results are averaged over these three environments (they have similar magnitudes and variance). Learning rates are kept constant, they affect the magnitude but not the shape of these curves. We use 10M steps in the control setting, and 500k steps otherwise.

\begin{figure}[ht]
\vskip 0.2in
\begin{center}
\centerline{\includegraphics[width=\columnwidth]{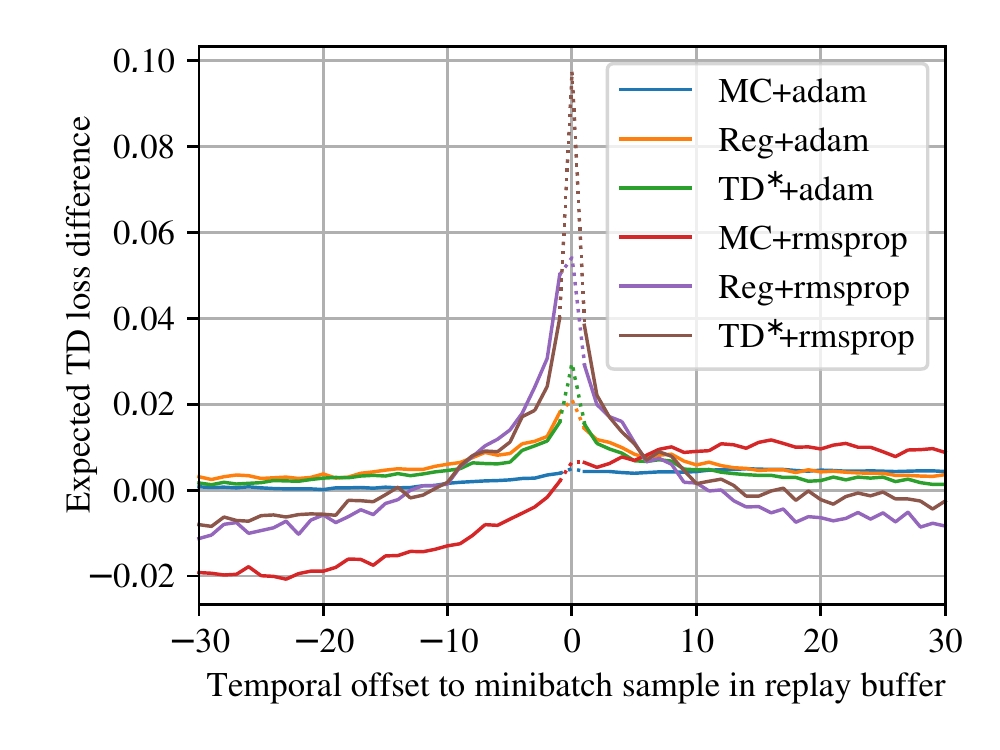}}
\caption{Regression on Atari: loss difference as a function of temporal offset in the replay buffer from the update sample. We use dotted lines at 0 offset to emphasize that the corresponding state was used for the update. The curve around 0 is indicative of the constructive interference of the TD and regression objectives.}
\label{reg_near_td_gain}
\end{center}
\vskip -0.2in
\end{figure}

We first use the following 3 supervised objectives to train models using $\mathcal{D}^*$ as a dataset and $Q_{\theta^*}$ as a \emph{distillation} target:
\begin{align*}
    \mathcal{L}_{MC}(s,a)&= (Q_\theta(s,a) - G^{(\mathcal{D}^*)}(s))^2 \label{eq:L_MC} \\
    \mathcal{L}_{reg}(s,a)&= (Q_\theta(s,a) - Q_{\theta^*}(s,a))^2 \\
    \mathcal{L}_{TD^*}(s,a,r,s')&= (Q_\theta(s,a) - (r + \gamma \max_{a'}Q_{\theta^*}(s',a')))^2
\end{align*}
and measure the difference in pointwise TD loss ($\mathcal{L}_{QL}$) for states \emph{surrounding} the state used for the update (i.e. states with a temporal offset of $\pm 30$ in the replay buffer trajectories), shown in Fig. \ref{reg_near_td_gain}.

There, we see that curves tend to be positive around $x=0$ (the sample used in the update), especially from indices -10 to 10, showing that \textbf{constructive interference is possible} when learning to approximate $Q^*$ with this data.
Since $Q_{\theta^*}$ is a good approximation, we expect that $Q_{\theta^*}(s,a) \approx (r + \gamma \max_{a'}Q_{\theta^*}(s',a'))$, so  $\mathcal{L}_{reg}$ and $\mathcal{L}_{TD^*}$ have similar targets and we expect them to have similar behaviours. Indeed, their curves mostly overlap. 

Next, we again measure the difference in pointwise loss for surrounding states. We train control agents and policy evaluation (or \emph{Batch Q}) agents with the Q-Learning loss:
\begin{equation}
    \mathcal{L}_{QL} = [Q_\theta(S_t,A_t) - (R_t + \gamma \max_{a}Q_{\theta}(S_{t+1},a))]^2.
\end{equation}
\begin{figure}[ht]
\vskip 0.2in
\begin{center}
\centerline{\includegraphics[width=\columnwidth]{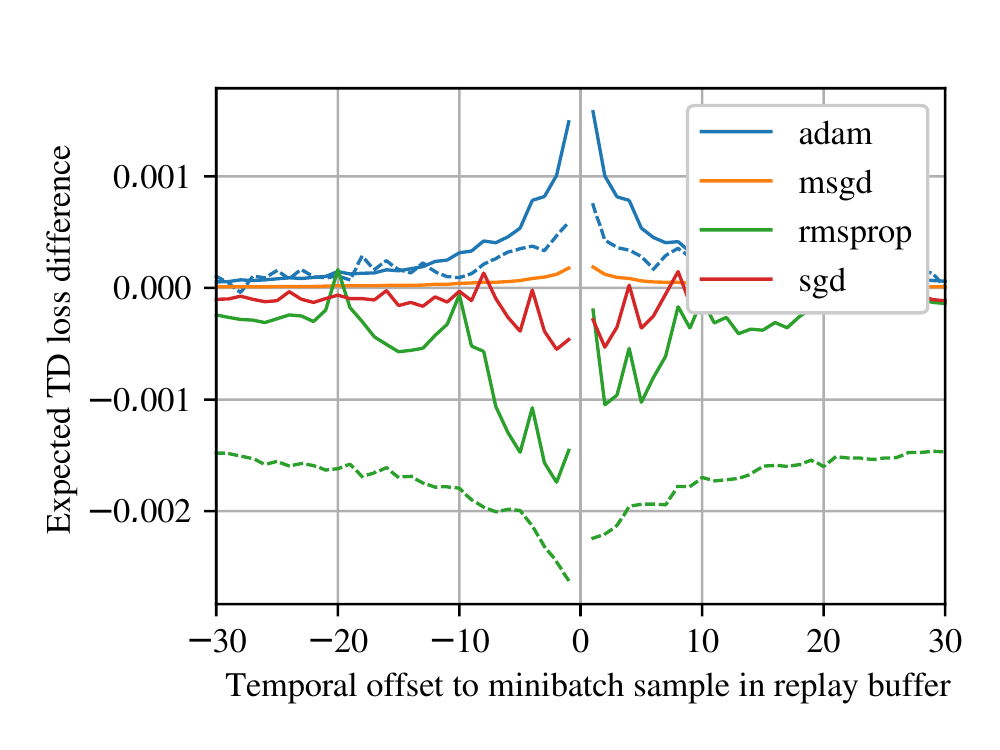}}
\caption{TD Learning on Atari: loss difference as a function of offset in the replay buffer of the update sample. Full lines represent Q-Learning control experiments, while dashed lines represent policy evaluation with a Q-Learning objective. We exclude $x=0$ for clarity, as it has a high value (see appendix Fig. \ref{ql_and_sarsa_td_gain_with_middle}). Compared to regression, the magnitude of the gain is much smaller. }
\label{ql_and_sarsa_td_gain}
\end{center}
\vskip -0.2in
\end{figure}
We show the results in Fig. \ref{ql_and_sarsa_td_gain}. Compared to the regressions in Fig. \ref{reg_near_td_gain}, the pointwise difference is more than an order of magnitude smaller, and drops off even faster when going away from $x=0$. This suggests a low interference, and a low update propagation. For certain optimizers, here RMSProp \citep{hinton2012neural} and SGD, this effect is even slightly negative. We believe this difference may be linked to momentum (note the difference with Adam \citep{kingma2015adam} and Momentum-SGD), which might dampen some of the negative effects of TD on interference (further discussed in section \ref{sec:interf_and_momentum}).

Interestingly, while Q-Learning does not have as strong a gain as the regressions from Fig. \ref{reg_near_td_gain}, it has a larger gain than policy evaluation. This may have several causes, and we investigate two. 

First, because of the initial random exploratory policy, the DNN initially sees little data variety, and may be able to capture a minimal set of factors of variation; then, upon seeing new states, the extracted features are forced to be mapped onto those factors of variation, improving them, leading to a natural curriculum. By looking at the \emph{singular values} of the last hidden layer's matrix after 100k steps, we do find that there is a \emph{consistently larger spread} in the policy evaluation case than the control case (see appendix \ref{sec:appendix:figures-singular-values}), showing that in the control case fewer factors are initially captured. This effect diminishes as training progresses.

Second, having run for 10M steps, control models could have been trained on more data and thus be forced to generalize better; this turns out \textbf{not} to be the case, as measuring the same quantities for only the first 500k steps yields very similar magnitudes. In other words, after a few initial epochs, function interference remains constant (see appendix \ref{sec:appendix:figures-of-evolution-of-td-gain}).

Interestingly, these results are consistent with those of \citet{agarwal2019striving}, who study off-policy learning. Among many other results, \citet{agarwal2019striving} find that off-policy-retraining a DQN model (i.e. Batch Q-Learning) on another DQN agent's lifetime set of trajectories yields much worse performance. This is consistent with our results showing more constructive interference in control than in policy evaluation, and suggests that the order in which data is presented may matter when bootstrapping is used.

\subsection[TD(lambda) and bootstrapping]{TD($\lambda$) and bootstrapping}

A central hypothesis of this work is that bootstrapping causes instability in interference, causing it to become small and causing models to memorize more. Here we perform policy evaluation on $\mathcal{D}^*$ with TD($\lambda$). TD($\lambda$) is by design a way to tradeoff between bias and variance in the target by trading off between few-step bootstrapped targets and long-term bootstrapped targets which approach Monte-Carlo returns. In other words, TD($\lambda$) allows us to diminish reliance on bootstrapping.

\begin{figure}[ht]
\vskip 0.2in
\begin{center}
\centerline{\includegraphics[width=\columnwidth]{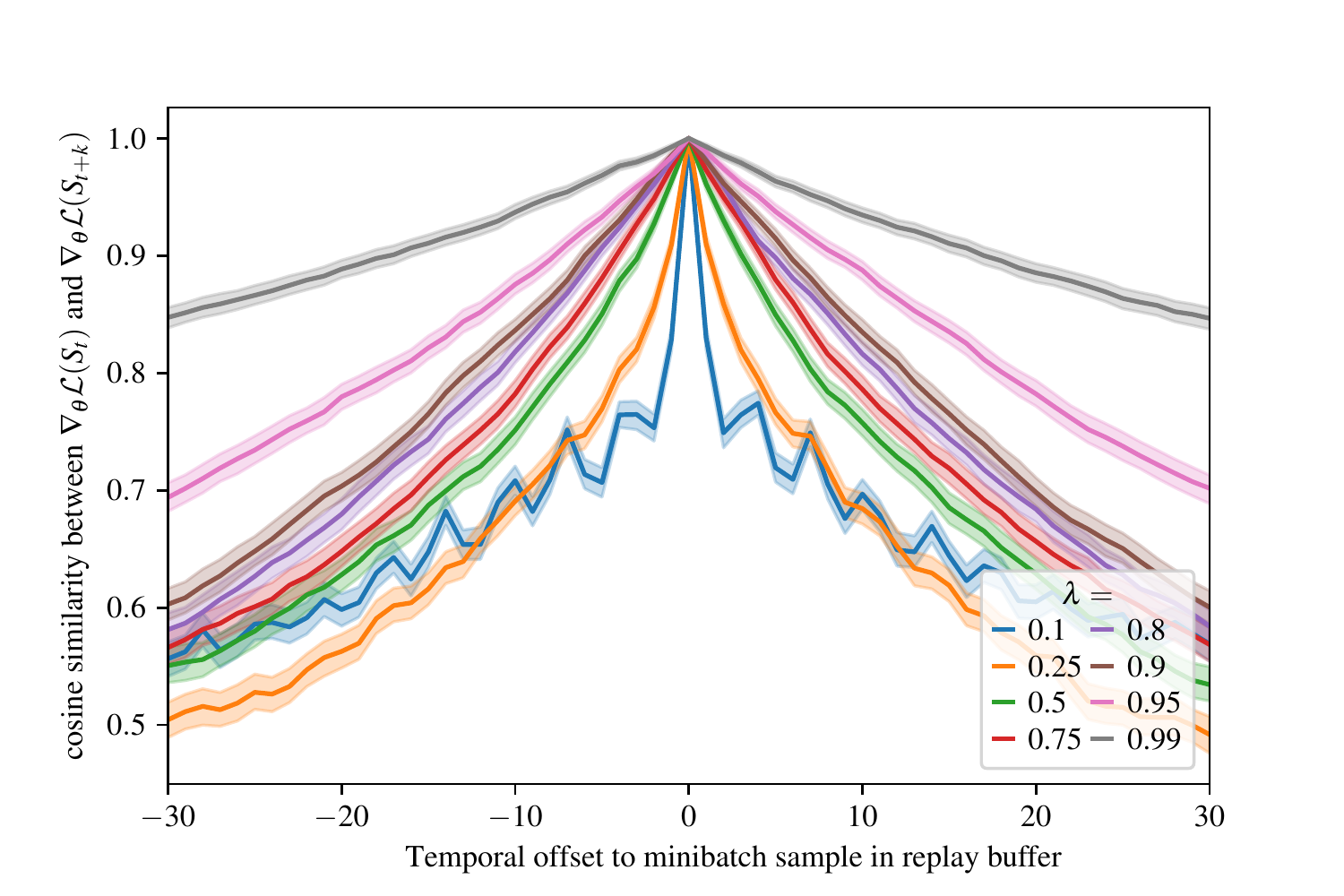}}
\caption{Cosine similarity between gradients at $S_t$ (offset $x=0$) and the gradients at the neighboring states in the replay buffer (MsPacman). As $\lambda$ increases, so does the temporal coherence of the gradients.}
\label{td_lambda_cossim}
\end{center}
\vskip -0.2in
\end{figure}

This tradeoff is especially manifest when measuring the \emph{stiffness} of gradients (cosine similarity) as a function of temporal offset, as shown in Fig. \ref{td_lambda_cossim}. There we see that the closer $\lambda$ is to 1, the more gradients are similar around an update sample, suggesting that diminishing reliance on bootstrapping reduces the effect of TD inducing low-interference memorizing parameterizations.

Note that this increase in similarity between gradients is also accompanied by an increase in pointwise loss difference (see appendix Fig. \ref{fig:lambda_td_near_td_gain}), surpassing that of Q-Learning (Fig. \ref{ql_and_sarsa_td_gain}) in magnitude. This suggests that TD($\lambda$) offers more coherent targets that allow models to learn faster, for sufficiently high values of $\lambda$.

\subsection{The high variance of bias in TD(0)}



In TD(0), the current target for any state depends on the prediction made at the next state. The difference between that prediction and the true value function makes the target a biased estimator when bootstrapping is in progress and information flows from newly visited states to seen states. 

This ``bootstrap bias'' itself depends on a function approximator which has its own bias-variance tradeoff (in the classical sense). For a high-variance approximator, this bootstrap bias might be inconsistent, making the value function alternate between being underestimated and being overestimated, which is problematic in particular for nearby states\footnote{Consider these two sequences of predictions of $V$: $[1,2,3,4,5,6]$ and $[1,2,1,2,1,2]$. Suppose no rewards, $\gamma=1$, and a function interference ($\bar{\rho}$) close to 1 for illustration, both these sequences have the same average TD(0) error, 1, yet the second sequence will cause any TD(0) update at one of the states to only correctly update half of the values.}. In such a case, a gradient descent procedure cannot truly take advantage of the constructive interference between gradients.

Indeed, recall that in the case of a regression, interference can be decomposed as:
\begin{align*}
    \rho_{x,y} = \frac{\partial J(x)}{\partial f(x)}\frac{\partial f(x)}{\partial \theta}^T\frac{\partial f(y)}{\partial \theta}\frac{\partial J(y)}{\partial f(y)},
\end{align*}
which for the TD error $\delta_x = V(x) - (r(x) + \gamma V(x'))$ with $x'$ some successor of $x$, can be rewritten as:
\begin{align*}
    \rho_{x,y} = \delta_x \delta_y \nabla_\theta V(x)^T \nabla_\theta V(y).
\end{align*}
If $x$ and $y$ are nearby states, in some smooth high-dimensional input space (e.g. Atari) they are likely to be close in input space and thus to have a positive function interference $\nabla_\theta V(x)^T \nabla_\theta V(y)$. If the signs of $\delta_x$ and $\delta_y$ are different, then an update at $x$ will increase the loss at $y$. As such, we measure the variance of the sign of the TD error along small windows (of length 5 here) in trajectories as a proxy of this local target incoherence.

\begin{figure}[ht]
\vskip 0.2in
\begin{center}
\centerline{\includegraphics[width=\columnwidth]{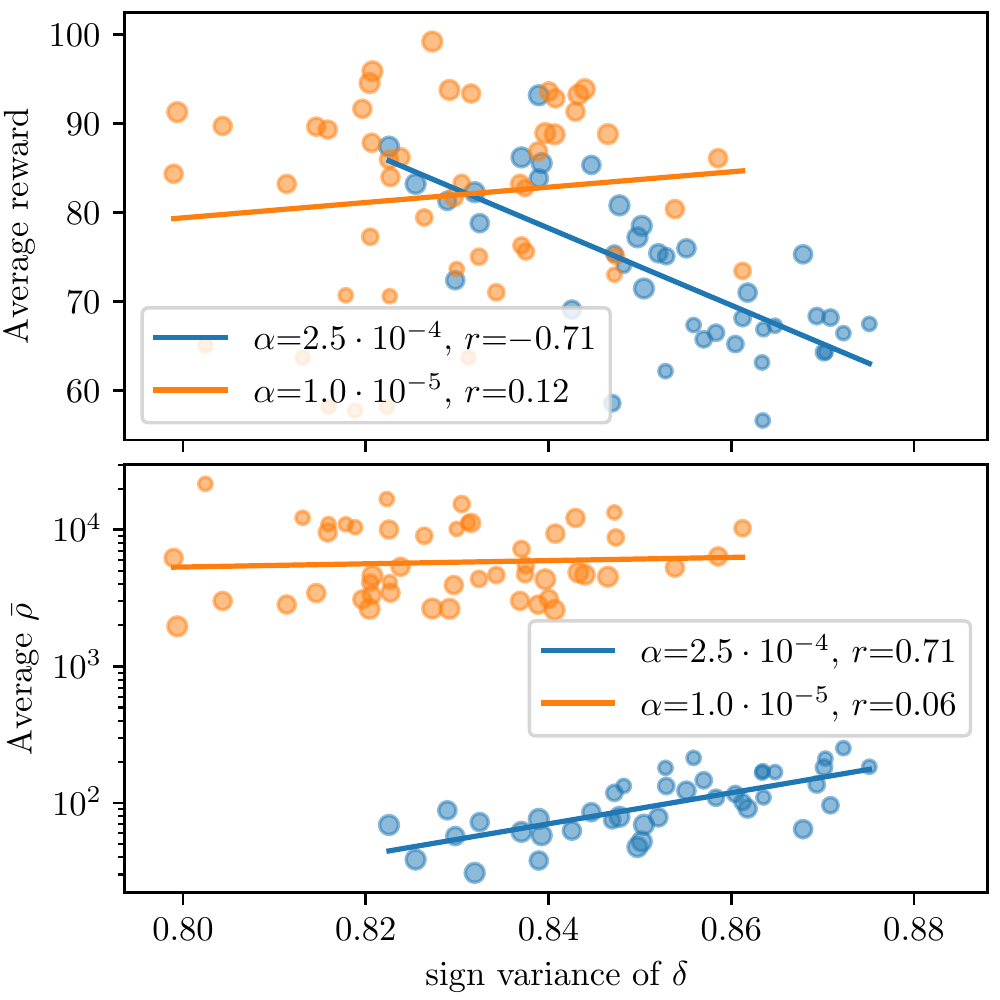}}
\caption{Top, average reward after training as a function of the sign variance for different learning rates ($\alpha$) and number of hidden units (size of markers). We can see that by using a much smaller learning rate than normal, the biasing effect of TD is lessened, at the cost of many more updates. Bottom, average function interference $\bar{\rho}$ after training. We see that, as predicted, parameterizations with large $\bar{\rho}$ and a large sign variance perform much worse (note that the $x$-axis of both plots are aligned, allowing for an easy reward/interference comparison).}
\label{var_vs_reward_mspacman}
\end{center}
\vskip -0.2in
\end{figure}

We observe this at play in Fig. \ref{var_vs_reward_mspacman}, which shows interference and rewards as a function of sign variance for a DQN agent trained on MsPacman. As predicted, parameterizations with a large $\bar{\rho}$ and a large sign variance perform much worse. We note that this effect can be lessened by using a much smaller learning rate than is normal, but this comes at the cost of having to perform more updates for a similar performance (in fact, presumably because of reduced instability, performance is slightly better, but only towards the end of training; runs with a normal $\alpha$ plateau halfway through training).

Interestingly, parameterizations with large $\bar{\rho}$ \emph{generally do} have a large sign variance ($r=0.71$) in the experiment of Fig. \ref{var_vs_reward_mspacman}. Indeed, we believe that the evolution of interference in TD methods may be linked to sign variance, the two compounding together, and may explain this trend.

These results are consistent with the improvements obtained by \citet{thodoroff2018temporal}, who force a temporal smoothing of the value function through convex combinations of sequences of values which likely reduces sign variance. These results are also consistent with \citet{anschel2017averaged} and \citet{agarwal2019striving} who obtain improvements by training ensembles of value functions. Such ensembles should partly reduce the sign variance of $\delta$, as bias due to initialization should average to a small value, making targets more temporally consistent.

Finally, note that in regression, this problem may eventually go away as parameters converge. Instead, in TD(0), especially when making use of a frozen target, this problem simply compounds with time and with every update. In what follows we consider this problem analytically.

\section{Understanding the evolution of interference}

Here we attempt to provide some insights into how interference evolves differently in classification, regression, and TD learning. For detailed derivations see appendix \ref{app:derivations}.

Recall that the interference $\rho$ can be obtained by the negative of the derivative of the loss $J(A)$ after some update using $B$ w.r.t. the learning rate $\alpha$, i.e. 
\begin{align}
    \theta' &= \theta - \alpha \nabla_\theta J(B)\\
    \rho_{AB} = -\partial J_{\theta'}(A)/\partial\alpha &= \nabla_{\theta'} J(A) \cdot \nabla_\theta J(B)\\
    &\approx \nabla_{\theta} J(A) \cdot \nabla_\theta J(B).
\end{align}
The last step being a simplification as $\theta \approx \theta'$.

To try to understand how this quantity evolves, we can simply take the derivative of $\rho$ (and $\bar\rho$) w.r.t. $\alpha$ but evaluated at $\theta'$, that is,  $\rho'_{AB} = \partial (\nabla_{\theta'} J(A) \cdot \nabla_{\theta'} J(B))/\partial \alpha$. In the general case, we obtain (assuming $\theta \approx \theta'$, we omit the $\theta$ subscript and subscript $A$ and $B$ for brevity):
\begin{align}
    \rho'_{AB} = -(\nabla J_B^T H_A + \nabla J_A^T H_B) \nabla J_B \label{eq:rho_prime_AB}\\
    \bar\rho'_{AB} = -(\nabla f_B^T \bar H_A + \nabla f_A^T \bar H_B) \nabla J_B \label{eq:bar_rho_prime_AB}
\end{align}
where $H_A = \nabla^2_\theta J(A;\theta)$, $\bar H_A = \nabla^2_\theta f(A;\theta)$ are Hessians.

Interpreting this quantity is non-trivial, but consider $\nabla f_A^T \bar H_B \nabla J_B$; parameters which make $f_A$ change, which have high curvature at $B$ (e.g. parameters that are not stuck in a saddle point or a minima at $B$), and which change the loss at $B$ will contribute to change $\rho$. Understanding the sign of this change requires a few more assumptions.

Because neural networks are somewhat smooth (they are Lipschitz continuous, although their Lipschitz constant might be very large, see \citet{scaman2018lipschitz}), it is likely for examples that are close in input space \emph{and} target space to have enough gradient and curvature similarities to increase their corresponding interference, while examples that are not similar would decrease their interference. Such an interpretation is compatible with our results, as well as those of \citet{fort2019stiffness} who find that \emph{stiffness} (cosine similarity of gradients) is mostly positive only for examples that are in the same class. 

Indeed, notice that for a given softmax prediction $\sigma$, for $A$ and $B$ of different classes $y_A,y_B$, the sign of the partial derivative at $\sigma_{y_A}(A)$ will be the opposite of that of $\sigma_{y_A}(B)$. Since gradients are multiplicative in neural networks, this will flip the sign of all corresponding gradients related to $\sigma_{y_A}$, causing a mismatch with curvature, and a decrease in interference. Thus the distribution of targets and the loss has a large role to play in aligning gradients, possibly just as much as the input space structure.

 We can also measure $\rho'$ to get an idea of its distribution. For a randomly initialized neural network, assuming a normally distributed input and loss, we find that it does not appear to be 0 mean. While the median is close to 0, but consistently negative, the distribution seems heavy-tailed with a slightly larger negative tail, making the negative mean further away from 0 than the median. In what follows we decompose $\rho'$ to get some additional insights.

In the case of regression, $J_A = 1/2(f_A - y_A)^2$, $\delta_A = (f_A - y_A)$, we get that:
\begin{align}
    \rho'_{reg;AB}& =
    -\bar{\rho}_{AB}^{2}\delta_{B}^{2}-2\delta_{A}\delta_{B}\bar{\rho}_{AB}\bar{\rho}_{BB} \nonumber\\
    &\phantom{{}=}-\delta_{A}\delta_{B}^{2}\nabla f_B (\bar H_A\nabla f_B+\bar H_B\nabla f_A) \label{eq:rhop-reg}
\end{align}
Another interesting quantity is the evolution of $\rho$ when $J$ is a TD loss if we assume that the bootstrap target also changes after a weight update. With the $\theta\approx\theta'$ simplification, $\delta_A=V_A-(r+\gamma V_{A'})$ the TD error at $A$, $A'$ some successor of $A$, we get:
\begin{align}
\rho'_{TD;AB}	&= 
    - \delta_{B}^{2}\bar{\rho}_{AB}(\bar{\rho}_{AB}-\gamma \bar{\rho}_{A'B})\nonumber\\
    &\phantom{{}=}-\delta_{A}\delta_{B}\bar{\rho}_{AB}(\bar{\rho}_{BB}-\gamma \bar{\rho}_{B'B})\nonumber\\
    &\phantom{{}=}-\delta_{A}\delta_{B}^{2}\nabla{f_B}(\bar{H}_A\nabla{f_B}+\bar{H}_B\nabla f_A) \label{eq:rhop-td}
\end{align}
Again considering the smoothness of neural networks, if $A$ and $B$ are similar, but happen to have opposite $\delta$ signs, their interference will decrease. Such a scenario is likely for high-capacity high-variance function approximators, and is possibly \textbf{compounded by the evolving loss landscape}. As the loss changes--both prediction and target depend on a changing $\theta$--it evolves imperfectly, and there is bound to be many pairs of nearby states where only one of the $\delta$s flips signs, causing gradient misalignments. This would be consistent with our finding that higher-capacity neural networks have a smaller interference in TD experiments (see appendix Fig. \ref{fig:interf_vs_capacity}) while the reverse is observed in classification. 

We now separately measure the three additive terms of $\rho'_{reg}$ and $\rho'_{TD}$, which we refer to as $\rho'=-r_1-r_2-r_3$, in the same order in which they appear in \eqref{eq:rhop-reg} and \eqref{eq:rhop-td}. 

\begin{figure}[ht]
\begin{center}
\centerline{\includegraphics[width=\columnwidth]{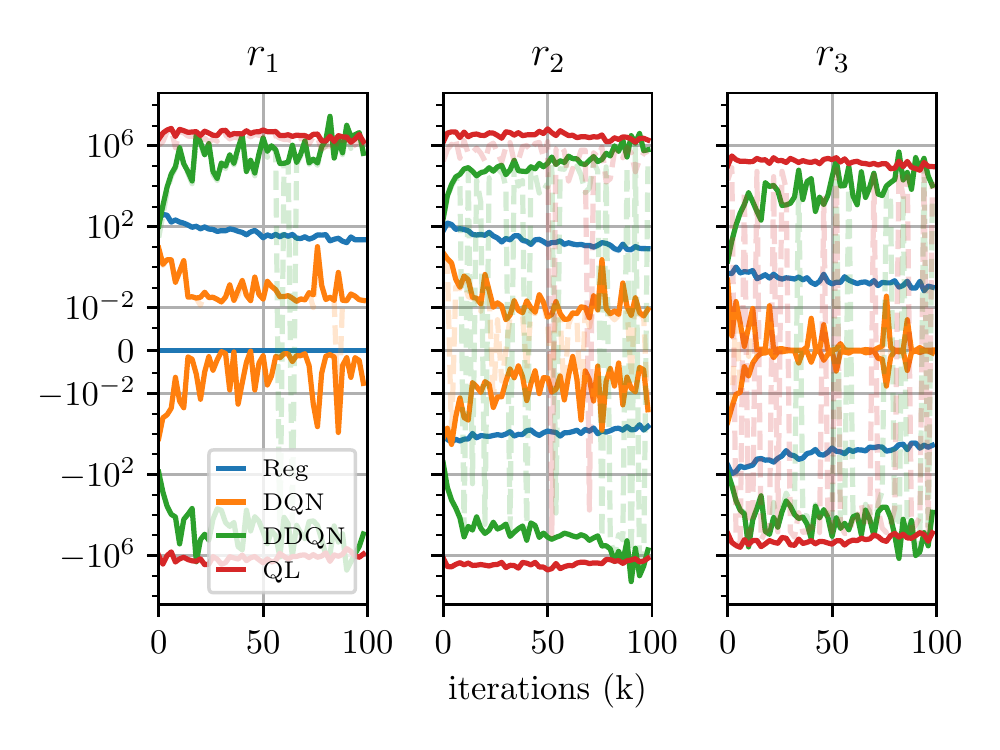}}
\caption{$r_1,r_2,r_3$ for $\rho'_{reg}$ (Reg) and $\rho'_{TD}$ (DQN, DDQN, QL) measured early in training. The transparent dashed lines are the mean $r_i$, averaged over 1024 ($32\times 32$) sample pairs, averaged over 3 runs. The full lines above and below 0 are the average of the positive and negative samples of $r_i$ respectively. These lines show the relative magnitudes of each part: in general, positive samples dominate for $r_1$, $r_2$ varies a lot between positive and negative for TD, while $r_3$ is mostly negative with some variance for TD.}
\label{fig:rhop_terms}
\end{center}
\end{figure}

We measure these terms in four scenarios, using a MsPacman expert replay buffer. We regress to $Q_{\theta*}$ (measuring $\rho'_{reg}$), and run policy evaluation with three different targets (measuring $\rho'_{TD}$). In DQN, the target $Q_{\bar\theta}$ is a frozen network updated every 10k iterations; in DDQN the target is updated with an exponential moving average rule, $\bar\theta = (1-\tau)\bar\theta +\tau\theta$, with $\tau=0.01$; in QL the target is the model itself $Q_\theta$ as assumed in \eqref{eq:rhop-td}. This is shown in Fig. \ref{fig:rhop_terms}. We see that in regression $r_1$ and $r_2$ are positive much more often than they are negative, while in TD methods, the positive samples tend to dominate but the proportion of negative samples is much larger, especially for $r_2$, which contains a $\delta_A\delta_B$ product. We see that $r_3$ tends to have a smaller magnitude than other terms for TD methods, and is negative on average. These results again suggest that TD methods do not have a stable evolution of interference.

\subsection{Interference and momentum}
\label{sec:interf_and_momentum}

Momentum SGD has the following updates, $\beta \in [0,1)$:
\begin{align}
    \mu_{t}&=(1-\beta)\nabla_{\theta}{J_{B}}+\beta\mu_{t-1}\\
    \theta' &= \theta - \alpha (\beta \mu_{t-1} + (1-\beta) \nabla_\theta J_B)
\end{align}
yielding the following quantities:
\begin{align}
    \rho_{\mu;AB}&=(1-\beta)\nabla_{\theta'}{J_{A}}\cdot\nabla_{\theta}{J_{B}}+\beta\nabla_{\theta'}{J_{A}}\cdot\mu_{t-1}\\
    \rho'_{\mu;AB} &= -(1-\beta)\rho'_{AB} - \beta \nabla J_{B}H_{A}\mu_{t-1}
\end{align}
Note that the first term of $\rho'_{\mu;AB}$ is simply eq. \eqref{eq:rho_prime_AB} times $1-\beta$. The second term is more interesting, and presumably larger as $\beta$ is usually close to 1. It indicates that for interference to change, the curvature at $A$ and the gradient at $B$ need to be aligned with $\mu$, the moving average of gradients. As such, the evolution of interference may be driven more by the random (due to the stochasticity of SGD) alignment of the gradients with $\mu$, which should be stable over time since $\mu$ changes slowly, than by the (high-variance) alignment of curvature at $A$ and gradient at $B$. As such, momentum should lower the variance of $\rho'$ and dampen the evolution of interference when it is high-variance, possibly including dampening the negative effects of interference in TD.

\vfill

\section{Discussion}

RL is generally considered  a harder problem than supervised learning due to the non-i.i.d. nature of the data. Hence, the fact that TD-style methods require more samples than supervised learning when used with neural networks is not necessarily surprising. However, with the same data and the same final targets (the ``true" value function), it is not clear why TD updates lead to parameters that generalize worse than in supervised learning. Indeed, our results show that the interference of a converged model evolves differently as a function of data and capacity in TD than in supervised learning.

Our results also show that Q-Learning generalizes poorly, leading to  DNNs that memorize the training data (not unlike table lookup). Our results also suggest that TD($\lambda$), although not widely used in recent DRL, improves generalization. Additionally, we find differences between Adam and RMSProp that we initially did not anticipate. Very little work has been done to understand and improve the coupling between optimizers and TD, and our results indicate that this would be an important future work direction.

While a full description of the mechanisms that cause TD methods to have such problems remains elusive, we find that understanding the evolution of gradient interference reveals intriguing differences between the supervised and temporal difference objectives, and hint at the importance of stable targets in bootstrapping. 


Finally, our work suggests that the RL community should pay special attention to the current research on generalization in DNNs, as naively approaching the TD bootstrapping mechanism as a supervised learning problem does not seem to leverage the full generalization potential of DNNs.


\section*{Acknowledgements}

The first author is grateful for funding from the  Fonds de recherche du Québec – Nature et technologies (FRQNT), as well great discussions from people at Deepmind and Mila. We are in particular grateful for feedback to Joshua Romoff, Harsh Satija, Veronica Chelu, Pierre Thodoroff, and  Yoshua Bengio. 

\bibliographystyle{icml2020}
\bibliography{main}

\clearpage
\onecolumn
\appendix 

\section{Derivations}
\label{app:derivations}
\subsection{Interference}

Consider an objective-based SGD update from $\nabla_\theta J$ using sample $B$ (here $A$ and $B$ can be understood as samples, but in general they can be tasks, or even entire data distributions):
$$\theta' = \theta - \alpha \nabla_\theta J(B)$$
The effect of this update on the objective elsewhere, here at sample $A$, can be understood as the derivative of the loss elsewhere with respect to the learning rate, yielding the well-known gradient interference quantity $\rho$:
\begin{align}
    \rho_{AB} = \frac{\partial J_{\theta'}(A)}{\partial\alpha} &= -\frac{\partial J_\theta(A)}{\partial\theta'} \frac{\partial\theta'}{\partial\alpha}\\
    &= -\nabla_{\theta'}J_{\theta'}(A)^T\nabla_\theta J_\theta(B)\\
    &\approx -\nabla_{\theta}J_{\theta}(A)^T\nabla_\theta J_\theta(B)
\end{align}

This quantity can also be obtained from the Taylor expansion of the loss difference at $A$ after an update at $B$:
\begin{align}
    J_{\theta'(B)}(A) - J_\theta(A) &\approx J_{\theta}(A) - J_\theta(A) + \nabla_\theta J(A)^T(\theta' - \theta) + O(||\theta'-\theta||^2)\\
&\approx -\alpha \nabla_{\theta}J_{\theta}(A)^T\nabla_\theta J_\theta(B)
\end{align}

For what follows we use the following notation for brevity: we subscript $f(A)$ as $f_A$, when writing $\nabla_{\theta'} J_A$ we imply that $J_A = J(A; \theta')$, when writing a gradient $\nabla$ or Hessian $H$ a lack of $\theta$ subscript implies $\nabla_\theta$ or $H_\theta$ rather than $\theta'$. 

\subsection{Second order quantities}

The derivative of $\rho$ w.r.t. $\alpha$, or second order derivative of $J_{\theta'}$ w.r.t. $\alpha$ is:
\begin{align}
    \frac{\partial^2 J_{\theta'}(A)}{\partial\alpha^2} &= -\frac{\partial}{\partial\alpha}\nabla_{\theta'}J_A^T\nabla_\theta J_B\\
    &= -(\frac{\partial (\nabla_{\theta'}J(A))}{\partial \theta'}\frac{\partial\theta'}{\partial\alpha})^T\nabla_\theta J_B\\
    &= -(-\nabla_{\theta'}^2 J_A \nabla_\theta J_B)^T \nabla_\theta J_B\\
    &\approx \nabla J_B^T H_A \nabla J_B
\end{align}

assuming $\theta\approx\theta'$ in the last step, and where $H_A = \nabla^2_\theta J_A$ is the Hessian. Again the only approximation here is $\theta\approx\theta'$

While this quantity is interesting, it is in a sense missing a part: what happens to the interference itself after an update? At 
\textbf{both} $A$ and $B$ at $\theta'$? 
\begin{align}
    \rho'_{AB} &= \frac{\partial}{\partial\alpha}\nabla_{\theta'}J_A^T\nabla_{\theta'} J_B\\
    &= (\frac{\partial (\nabla_{\theta'}J_A)}{\partial \theta'}\frac{\partial\theta'}{\partial\alpha})^T\nabla_\theta J_B
      + \nabla_{\theta'}J_A^T(\frac{\partial (\nabla_{\theta'}J_B)}{\partial \theta'}\frac{\partial\theta'}{\partial\alpha})
    \\
    &= (-\nabla_{\theta'}^2 J_A \nabla_\theta J_B)^T \nabla_\theta J_B + \nabla_{\theta'}J_A^T(-\nabla_{\theta'}^2 J_B \nabla_\theta J_B)\\
    &\approx -\nabla J_B^T H_A \nabla J_B - \nabla J_A^T H_B \nabla J_B
\end{align}

Following \citet{nichol2018first} we can rewrite this as:
\begin{align}
    =& -(\nabla J_B^T H_A + \nabla J_A^T H_B) \nabla J_B\\
    =& -(\nabla_\theta (\nabla J_B^T \nabla J_A)) \nabla J_B
\end{align}
This last form is easy to compute with an automatic differentiation software and does not require explicitly computing the hessian. We also verify empirically that this quantity holds with commonly used small step-sizes.

The derivative of function interference can also be written similarly:
\begin{align}
    \bar\rho'_{AB} &= \frac{\partial}{\partial\alpha}\nabla_{\theta'}f_A^T\nabla_{\theta'} f_B\\
    &= (\frac{\partial (\nabla_{\theta'}f_A)}{\partial \theta'}\frac{\partial\theta'}{\partial\alpha})^T\nabla_{\theta'} f_B
      + \nabla_{\theta'}f_A^T(\frac{\partial (\nabla_{\theta'}f_B)}{\partial \theta'}\frac{\partial\theta'}{\partial\alpha})
    \\
    &= (-\nabla_{\theta'}^2 f_A \nabla_\theta J(B))^T \nabla_{\theta'} f_B + \nabla_{\theta'}f_A^T(-\nabla_{\theta'}^2 f_B \nabla_\theta J(B))\\
    &\approx -\nabla J_B^T \bar{H}_A \nabla f_B - \nabla f_A^T \bar{H}_B \nabla J_B\\
    & = - (\nabla f_B^T \bar{H}_A + \nabla f_A^T \bar{H}_B) \nabla J_B
\end{align}
where by $\bar{H}$ we denote the Hessian of the function $f$ itself rather than of its loss.

Note that for the parameterized function $f_\theta$
$$\nabla_\theta J = \frac{\partial J}{\partial f}\frac{\partial f}{\partial \theta}$$

Let's write $\frac{\partial J}{\partial f}=\delta$. For any regression-like objective $(f-y)^2/2$, $\delta=(f-y)$. $\delta$'s sign will be positive if $f$ needs to decrease, and negative if $f$ needs to increase.

Let's rewrite the interference as:
$$ \nabla_{\theta}J_{\theta}(A)^T\nabla_\theta J_\theta(B) = \delta_A\delta_B \nabla_{\theta}f_{\theta}(A)^T\nabla_\theta f_\theta(B)$$

Then notice that $\rho'$ can be decomposed as follows. Let $g_{AB}=\nabla_{\theta}f_{\theta}(A)^{T}\nabla_{\theta}f_{\theta}(B)$
 , $g'_{AB}=\nabla_{\theta'}f_{\theta'}(A)^{T}\nabla_{\theta'}f_{\theta'}(B)$
 :
\begin{align}
\rho'_{reg;AB}= & \frac{\partial}{\partial\alpha}\delta_{A}\delta_{B}\nabla_{\theta'}f_{A}^{T}\nabla_{\theta'}f_{B}\\
= & \frac{\partial\delta_{A}}{\partial\alpha}\delta_{B}g'_{AB}+\frac{\partial\delta_{B}}{\partial\alpha}\delta_{A}g'_{AB}\\
 & +\delta_{A}\delta_{B}(\frac{\partial}{\partial\alpha}\nabla_{\theta'}f_{A})^{T}\nabla_{\theta'}f_{B}+\delta_{A}\delta_{B}(\frac{\partial}{\partial\alpha}\nabla_{\theta'}f_{B})^{T}\nabla_{\theta'}f_{A}\\
= & -\nabla_{\theta'}f_{A}^{T}\nabla_{\theta}J_{B}\delta_{B}g'_{AB}-\nabla_{\theta'}f_{B}^{T}\nabla_{\theta}J_{B}\delta_{A}g'_{AB}\\
 & +\delta_{A}\delta_{B}(-\bar{H}_{\theta';A}\nabla_{\theta}J_{B})^{T}\nabla_{\theta'}f_{B}+\delta_{A}\delta_{B}(-\bar{H}_{\theta';B}\nabla_{\theta}J_{B})^{T}\nabla_{\theta'}f_{A}\\
 & \mbox{if we assume \ensuremath{\theta\approx\theta'}, \ensuremath{g\approx g'} we can simplify}\\
\approx & -g\delta_{B}\delta_{B}g-2\delta_{B}g\delta_{A}g_{BB}-\delta_{A}\delta_{B}\delta_{B}\nabla_{\theta}f_{B}\bar{H}_{A}\nabla_{\theta}f_{B}-\delta_{A}\delta_{B}\delta_{B}\nabla_{\theta}f_{B}\bar{H}_{B}\nabla_{\theta}f_{A}\\
= & -g_{AB}^{2}\delta_{B}^{2}-2\delta_{A}\delta_{B}g_{AB}g_{BB}-\delta_{A}\delta_{B}^{2}\nabla_{\theta}f_{B}(\bar{H}_{A}\nabla f_{B}+\bar{H}_{B}\nabla_{\theta}f_{A})
\end{align}
We can also compute $\rho'$ for TD(0) assuming that the target is not frozen and is influenced by an update to $\theta$. Again we want $\partial/\partial\alpha[g'_{AB}]$ for an update at $B$, interference at $A$, assuming that $B'$ is 
a successor state of $B$ used for the TD update, and $A'$ a successor 
of $A$ in $\delta_{A}$:
\begin{align}\theta' & =\theta-\alpha\delta_{B}\nabla_{\theta}{f_B}\\
 & =\theta-\alpha(f_{B}-(r+\gamma f_{B'})\nabla_{\theta}f_B
\end{align}
Also note that:
\begin{align}
\frac{\partial\delta_{A}}{\partial\alpha}= & \left(\frac{\partial f_A}{\partial\theta'}\frac{\partial\theta'}{\partial\alpha}-\gamma\frac{\partial f_{A'}}{\partial\theta'}\frac{\partial\theta'}{\partial\alpha}\right)\\
= & -\delta_{B}(\nabla_{\theta'}{f_A}^T\nabla_{\theta}{f_B}-\gamma\nabla_{\theta'}{f_{A'}}^T\nabla_{\theta}{f_B})\\
\end{align}
Let $g_{AB}=\nabla_{\theta}f_A^{T}\nabla_{\theta}f_B$,
$g'_{AB}=\nabla_{\theta'}f_A^{T}\nabla_{\theta'}f_B$
and $g_{AB}^{\backslash}=\nabla_{\theta'}f_A^{T}\nabla_{\theta}f_B$:
\begin{align}
\rho_{TD;AB}&= \frac{\partial}{\partial\alpha}\left[\delta_{A}\delta_{B}\nabla_{\theta'}f_A^{T}\nabla_{\theta'}f_B\right]\\
&=  \frac{\partial}{\partial\alpha}\delta_{A}\delta_{B}\nabla_{\theta'}f_A^{T}\nabla_{\theta'}f_B\\
 &\phantom{{}=} +\delta_{A}\frac{\partial}{\partial\alpha}\delta_{B}\nabla_{\theta'}f_A^{T}\nabla_{\theta'}f_B\\
 &\phantom{{}=} +\delta_{A}\delta_{B}\frac{\partial}{\partial\alpha}\nabla_{\theta'}f_A^{T}\nabla_{\theta'}f_B\\
 &\phantom{{}=} +\delta_{A}\delta_{B}\nabla_{\theta'}f_A^{T}\frac{\partial}{\partial\alpha}\nabla_{\theta'}f_B\\
&=  -\delta_{B}(g_{AB}^{\backslash}-\gamma g_{A'B}^{\backslash})\delta_{B}g'_{AB}\\
 &\phantom{{}=} -\delta_{B}(g_{BB}^{\backslash}-\gamma g_{B'B}^{\backslash})\delta_{A}g'_{AB}\\
 &\phantom{{}=} +\delta_{A}\delta_{B}(-\bar H_{\theta';A}\nabla_{\theta}{J_B})^{T}\nabla_{\theta'}{f_B}+\delta_{A}\delta_{B}(-\bar H_{\theta';B}\nabla_{\theta}{J_B})^{T}\nabla_{\theta'}{f_A}
\end{align}
which again if we assume $\theta'\approx\theta,g_{AB}\approx g'_{AB}\approx g_{AB}^{\backslash}$, we can simplify to:

\begin{eqnarray*}
\rho'_{TD;AB} & = & -\delta_{B}^{2}g_{AB}(g_{AB}-\gamma g_{A'B})-\delta_{A}\delta_{B}g_{AB}(g_{BB}-\gamma g_{B'B})\\
 &  & -\delta_{A}\delta_{B}^{2}\nabla_{\theta} {f_B}(\bar H_A\nabla_{\theta}{f_B}+\bar H_B\nabla_{\theta}{f_A})
\end{eqnarray*}

\section{Architectures, hyperparameter ranges, and other experimental details}

We use the PyTorch library \citep{pytorch2019} for all experiments. To efficiently compute gradients for a large quantity of examples at a time we use the backpack library \citep{dangel2020backpack}.

Note that we run natural images experiments first to get a more accurate comparison of the generalization gap between RL and SL (Section 4). We then run Atari experiments to analyse information propagation, TD($\lambda$), and the local coherence of targets (Section 5), because Atari agents (1) have long term decision making which highlights the issues of using TD for long term reward predictions (which is TD's purpose) and (2) are a standard benchmark.

\label{app:hyperparams}

\subsection{Figure 1, 2, 8 and 9}

In order to generate these figures we train classifiers, regression models, DDQN agents and REINFORCE agents.

Models trained on SVHN and CIFAR10, either for SL, DDQN, or REINFORCE, use a convolutional architecture. Let $n_h$ be the number of hiddens and $n_L$ the number of extra layers. The layers are:
\begin{itemize}[leftmargin=1cm,topsep=0pt,itemsep=0pt,partopsep=0pt, parsep=0pt]
    \item Convolution, 3 in, $n_h$ out, filter size 5, stride 2
    \item Convolution, $n_h$ in, $2n_h$ out, filter size 3
    \item Convolution, $2n_h$ in, $4n_h$ out, filter size 3
    \item $n_L$ layers of Convolution, $4n_h$ in, $4n_h$ out, filter size 3, padding 1
    \item Linear, $4n_h\times10\times10$ in, $4n_h$ out
    \item Linear, $4n_h$ in, $n_o$ out.
\end{itemize}
All layers except the last use a Leaky ReLU \citep{maas2013rectifier} activation with slope 0.01 (note that we ran a few experiments with ReLU and tanh activations out of curiosity, except for the slightly worse training performance the interference dynamics remained fairly similar). For classifiers $n_o$ is 10, the number of classes. For agents $n_o$ is 10+4, since there are 10 classes and 4 movement actions.

Models trained on the California Housing dataset have 4 fully-connected layers: 8 inputs, 3 Leaky ReLU hidden layers with $n_h$ hiddens, and a linear output layer with a single output.

Models trained on the SARCOS dataset have 2+$n_L$ fully-connected layers: 21 inputs, 1+$n_L$ Leaky ReLU hidden layers with $n_h$ hiddens, and a linear output layer with 8 outputs.

Let $n_T$ be the number of training seeds. We use the following hyperparameter settings:
\begin{itemize}[leftmargin=1cm,topsep=0pt,itemsep=0pt,partopsep=0pt, parsep=0pt]
    \item SVHN, $n_h\in\{8,16,32\}$, $n_L\in\{0,1,2,3\}$, $n_T\in\{20, 100, 250, 500, 1000, 5000, 10000, 50000\}$
    \item CIFAR10, $n_h\in\{16,32,64\}$, $n_L\in\{0,1,2,3\}$, $n_T\in\{20, 100, 250, 500, 1000, 5000, 10000, 50000\}$
    \item SARCOS, $n_h\in\{16,32,64,128,256\}$, $n_L\in\{0,1,2,3\}$, $n_T\in\{20, 100, 250, 500, 1000, 5000, 10000, 44484\}$
    \item California Housing, $n_h\in\{16,32,64,128\}$, $n_T\in\{20, 100, 250, 500, 1000, 5000, 10000\}$
\end{itemize}
For SVHN and CIFAR10, we use the same architecture and hyperparameter ranges for classification, DDQN and REINFORCE experiments. Each hyperparameter setting is run with 3 or more seeds. The seeds affect the initial parameters, the sampling of minibatches, and the sampling of $\epsilon$-greedy actions.

Note that while we run REINFORCE on SVHN and CIFAR, we do not spend a lot of time analyzing its results, due the relatively low relevance of PG methods to the current work. Indeed, the goal was only to highlight the difference in trends between TD and PG, which do indicate that the two have different behaviours. Policy gradient methods do sometimes rely on the TD mechanism (e.g. in Actor-Critic), but they use different update mechanisms and deserve their own independent analysis, see for example \citet{ilyas2018deep}.

For optimizers, we use the standard settings of PyTorch:
\begin{itemize}[leftmargin=1cm,topsep=0pt,itemsep=0pt,partopsep=0pt, parsep=0pt]
    \item Adam, $\beta=(0.9, 0.999)$, $\epsilon=10^{-8}$
    \item RMSProp, $\alpha=0.99$, $\epsilon=10^{-8}$
    \item Momentum SGD, $\beta=0.9$ (with Nesterov momentum off)
\end{itemize}

\subsection{Figure 3, 4, 10, 12, and 13}

Figure \ref{reg_near_td_gain} is obtained by training models for 500k steps with a standard DQN architecture \citep{mnih2013playing}: 3 convolutional layers with kernels of shape $4\times 32 \times 8 \times 8$, $32\times 64 \times 4 \times 4$, and $64\times 64 \times 3 \times 3$ and with stride 4, 2, and 1 respectively, followed by two fully-connected layers of shape $9216 \times 512$ and $512 \times |\mathcal{A}|$, $\mathcal{A}$ being the legal action set for a given game. All activation are leaky ReLUs except for the last layer which is linear (as it outputs value functions). Experiments are run on MsPacman, Asterix and Seaquest for 10 runs each. A learning rate of $10^{-4}$ is used, with L2 weight regularization of $10^{-4}$.  We use $\gamma=0.99$, a minibatch size of 32, an $\epsilon$ of 5\% to generate $\mathcal{D}^*$, and a buffer size of 500k. The random seeds affect the generation of $\mathcal{D}^*$, the weight initialization, the minibatch sampling, and the choice of actions in $\epsilon$-greedy rollouts.

As per the previous figure, for Figure \ref{ql_and_sarsa_td_gain} we run experiments with a standard DQN architecture, train our policy evaluation models for 500k and our control models for 10M steps. When boostrapping to a frozen network, the frozen network is updated every 10k updates.

Figures 10, 12, and 13 also use results from these experiments.

\subsection{Figure 5, 11, and 13}

The experiments of Figure \ref{td_lambda_cossim} are run for 500k steps, as previously described, on MsPacman. $\lambda$-targets are computed with the forward view, using the frozen network to compute the target values -- this allows us to cheaply recompute all $\lambda$-targets once every 10k steps when we update the frozen network. Each setting is run with 5 random seeds.

Figures 11 and 13 also use results from these experiments.

\subsection{Figure 6}

Figure 6 reuses the results of Figure 4's policy evaluation experiments run with Adam.

\subsection{Figure 7}

Figure 7 uses the same experiment setup as in the Atari regression experiments on MsPacman, as well as policy evaluation experiments on MsPacman as previously described, all the while measuring individual terms of $\rho'_{reg}$ and $\rho'_TD$. Experiments are only run for the first 100k steps. Minibatches of size 32 are used.

\newpage

\section{Reproducibility checklist}

We follow the Machine Learning reproducibility checklist \citep{pineau2019checklist}, and refer to corresponding sections in the text when relevant.

For all models and algorithms presented, check if you include:

\begin{itemize}[leftmargin=1cm,topsep=0pt,itemsep=0pt,partopsep=0pt, parsep=0pt]
    \item \textbf{A clear description of the mathematical setting, algorithm, and/or model.} We use unmodified algorithms, described in the technical background, and only analyse their behaviour. The measures we propose are straightforward to implement and only require minimal changes. For more details see section \ref{app:hyperparams}.
     \item \textbf{An analysis of the complexity (time, space, sample size) of any algorithm.} The measures we propose only add a constant instrumentation overhead.
     \item \textbf{A link to a downloadable source code, with specification of all dependencies, including external libraries.} All code is included in supplementary materials, dependencies are documented within.
\end{itemize}

For any theoretical claim, check if you include:
\begin{itemize}[leftmargin=1cm,topsep=0pt,itemsep=0pt,partopsep=0pt, parsep=0pt]
    \item \textbf{A statement of the result.} See section \ref{app:derivations}.
    \item \textbf{A clear explanation of any assumptions.} idem.
    \item \textbf{A complete proof of the claim.} idem.
\end{itemize}

\begin{figure}[h]
\vskip 0.2in
\begin{center}
\centerline{\includegraphics[width=.5\columnwidth]{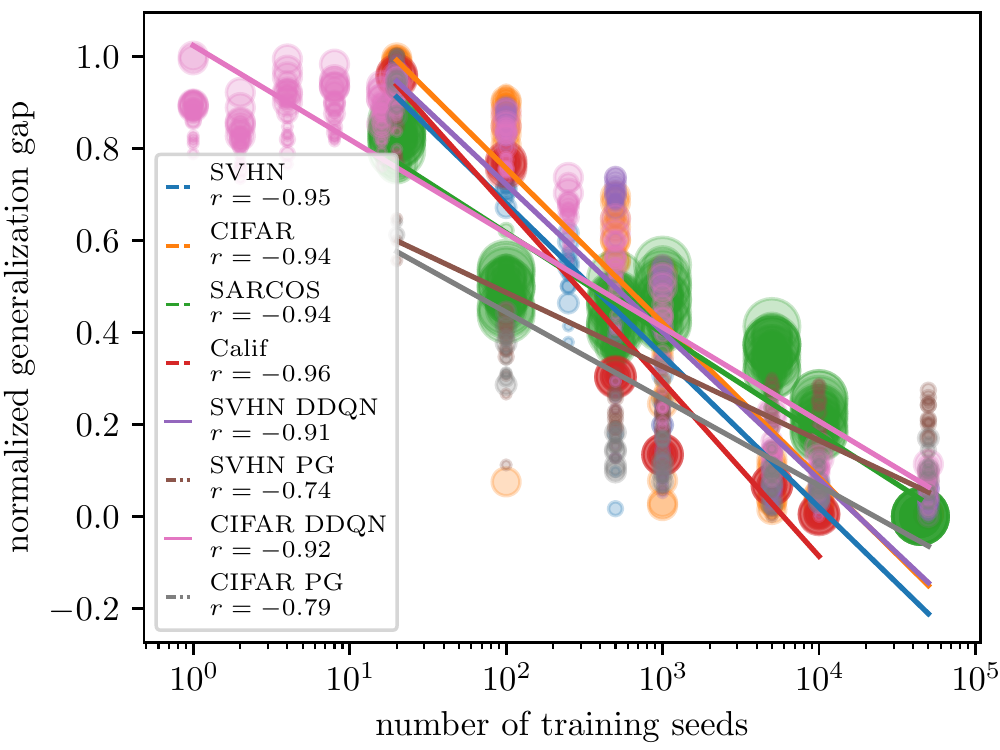}}
\caption{Generalization gap vs number of training seeds. The size of each circle (which represents a single experiment) is proportional to the number of hidden units.}
\label{gen_gap_vs_ntrain}
\end{center}
\vskip -0.2in
\end{figure}

For all figures and tables that present empirical results, check if you include:

\begin{itemize}[leftmargin=1cm,topsep=0pt,itemsep=0pt,partopsep=0pt, parsep=0pt]
    \item \textbf{A complete description of the data collection process, including sample size.} We collect data by running standard implementations of common algorithms with repeated runs.
    \item \textbf{A link to a downloadable version of the dataset or simulation environment.} Included in the code available in supplementary materials.
    \item \textbf{An explanation of any data that were excluded, description of any pre-processing step.} We generally chose hyperparameters that best represent state-of-the-art usage, then if necessary that best represent our findings. In most cases only minor learning rate adjutments were necessary, although they would not significantly change most plots.
    \item \textbf{An explanation of how samples were allocated for training / validation / testing.} We use standard train/valid/test splits as per the literature of each dataset.
    \item \textbf{The range of hyper-parameters considered, method to select the best hyper-parameter configuration, and specification of all hyper-parameters used to generate results.} See section \ref{app:hyperparams}.
    \item \textbf{The exact number of evaluation runs.} idem.
    \item \textbf{A description of how experiments were run.} idem.
    \item \textbf{A clear definition of the specific measure or statistics used to report results.} See section \ref{sec:comp-interf}.
    \item \textbf{Clearly defined error bars.} Figures with error bars compute a bootstrapped 90\% or 95\% confidence interval of the mean. We only use 90\% for Figure \ref{gen_gap_vs_ntrain} because of the many outliers.
    \item \textbf{A description of results with central tendency(e.g. mean) \& variation(e.g. stddev).} idem.
    \item \textbf{A description of the computing infrastructure used}     Almost all experiments were run on P100 and V100 GPUs, otherwise they were run on Intel i7 processors.

\end{itemize}

\newpage
\section{Extra figures}

\begin{figure}[h]
    \centering
    \includegraphics[width=.6\columnwidth]{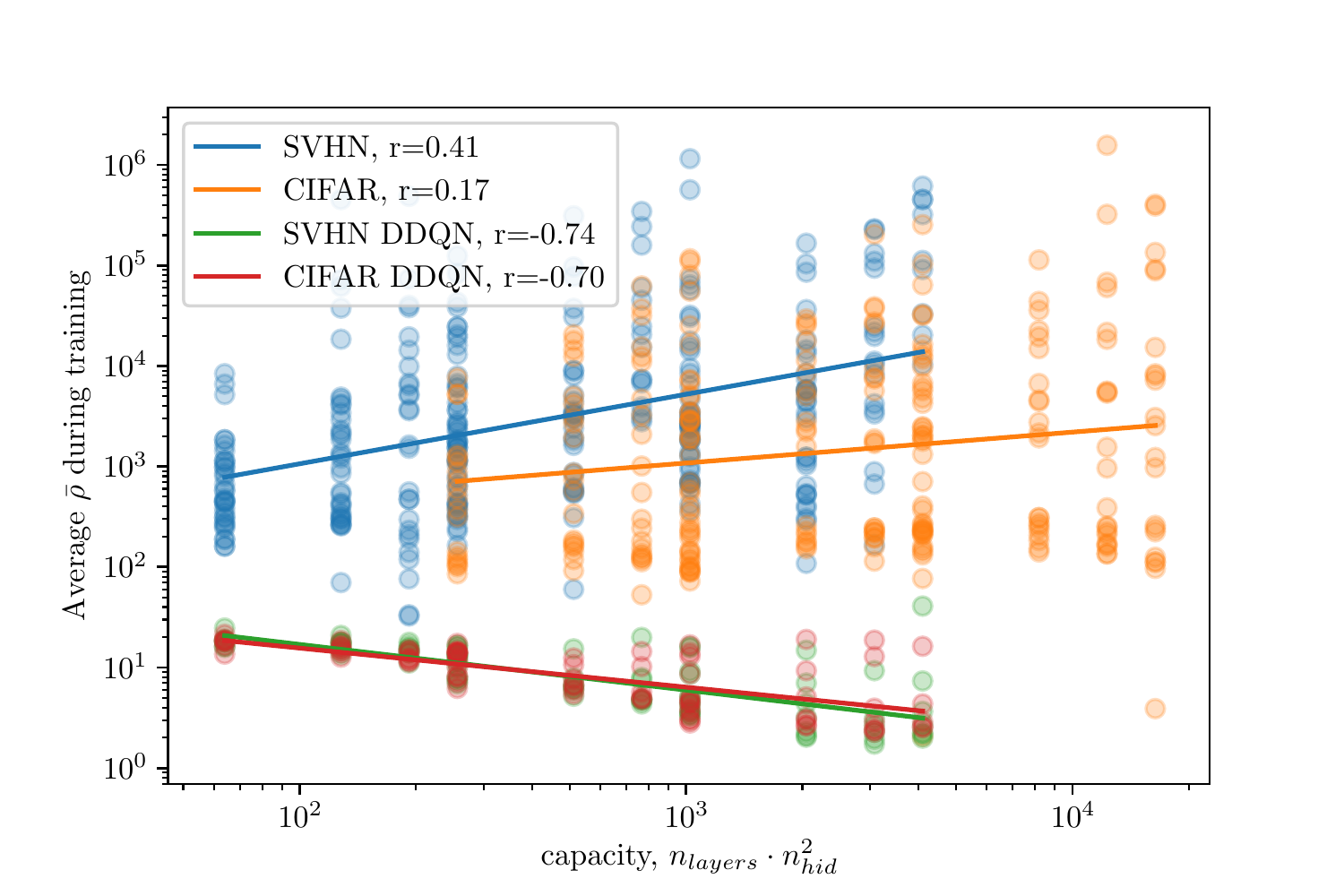}
    \caption{Average function interference during training as a function of capacity. TD methods and classifiers have very different trends.}
    \label{fig:interf_vs_capacity}
\end{figure}
\FloatBarrier

\begin{figure}[h]
\vskip 0.2in
\begin{center}
\centerline{\includegraphics[width=.54\columnwidth]{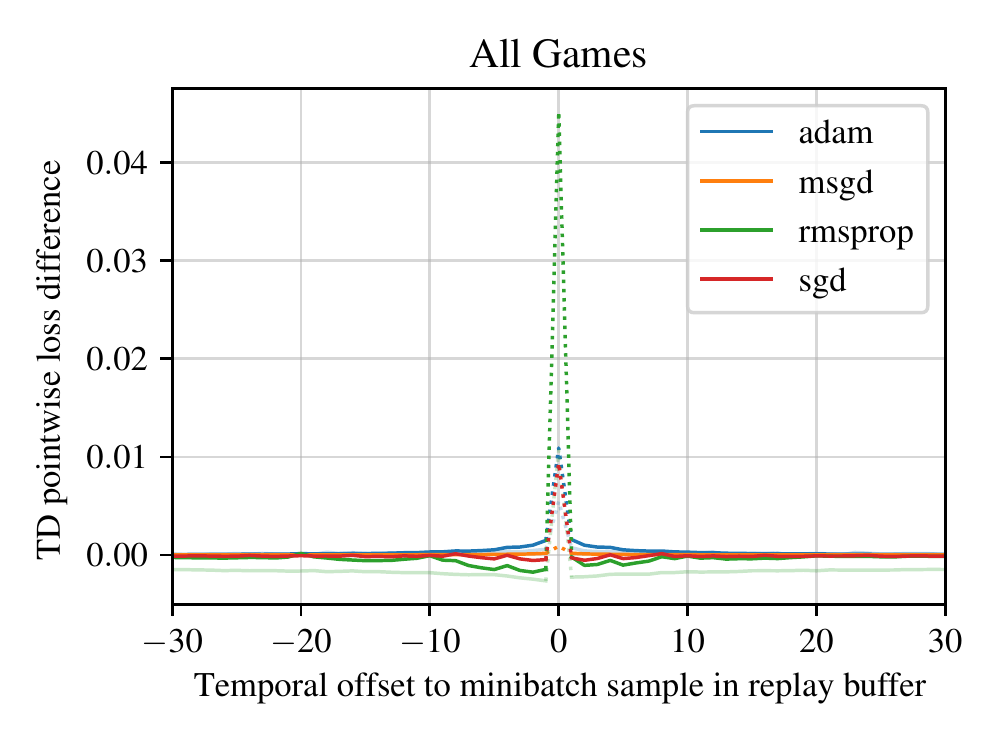}}
\caption{Reproduction of Figure \ref{ql_and_sarsa_td_gain} including $x=0$. RMSprop has a surprisingly large expected gain at $x=0$, but a negative gain around $x=0$, suggesting that RMSprop enables memorization more than Adam.}
\label{ql_and_sarsa_td_gain_with_middle}
\end{center}
\vskip -0.2in
\end{figure}
\FloatBarrier

\begin{figure}[h]
    \centering
    \includegraphics[width=.5\columnwidth]{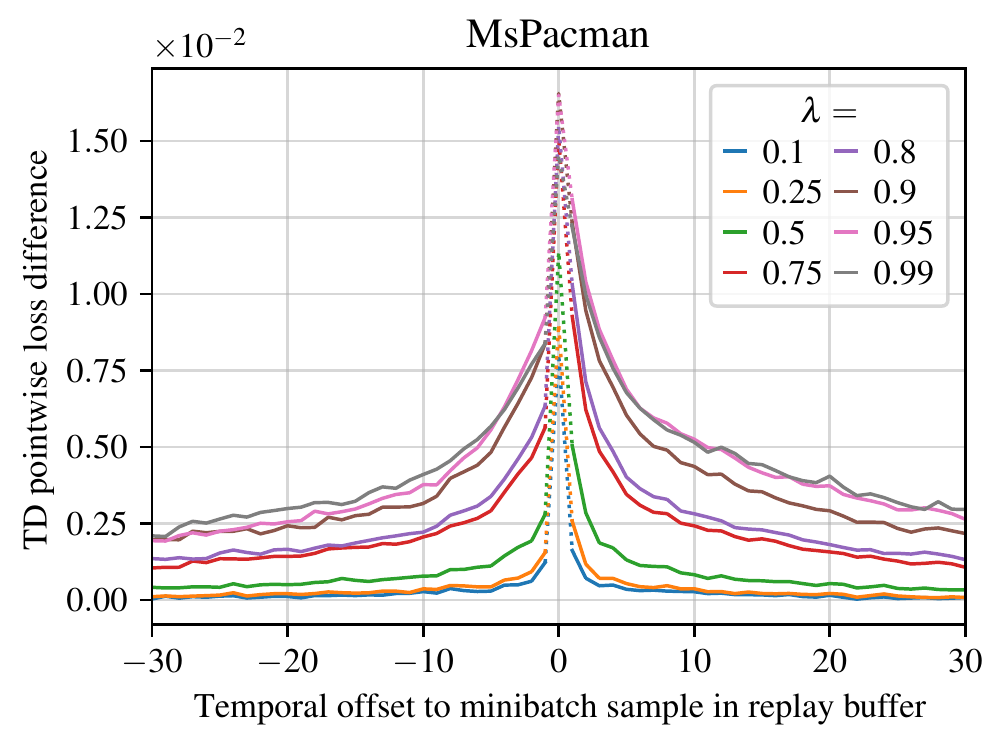}
    \caption{Evolution of TD pointwise loss difference, as a function of $\lambda$ in TD($\lambda$). Notice the asymmetry around 0.}
    \label{fig:lambda_td_near_td_gain}
\end{figure}
\FloatBarrier
%

\subsection{Singular values, control vs policy evaluation}
\label{sec:appendix:figures-singular-values}

\begin{figure}[h]
    \centering
    \includegraphics[width=0.6\textwidth]{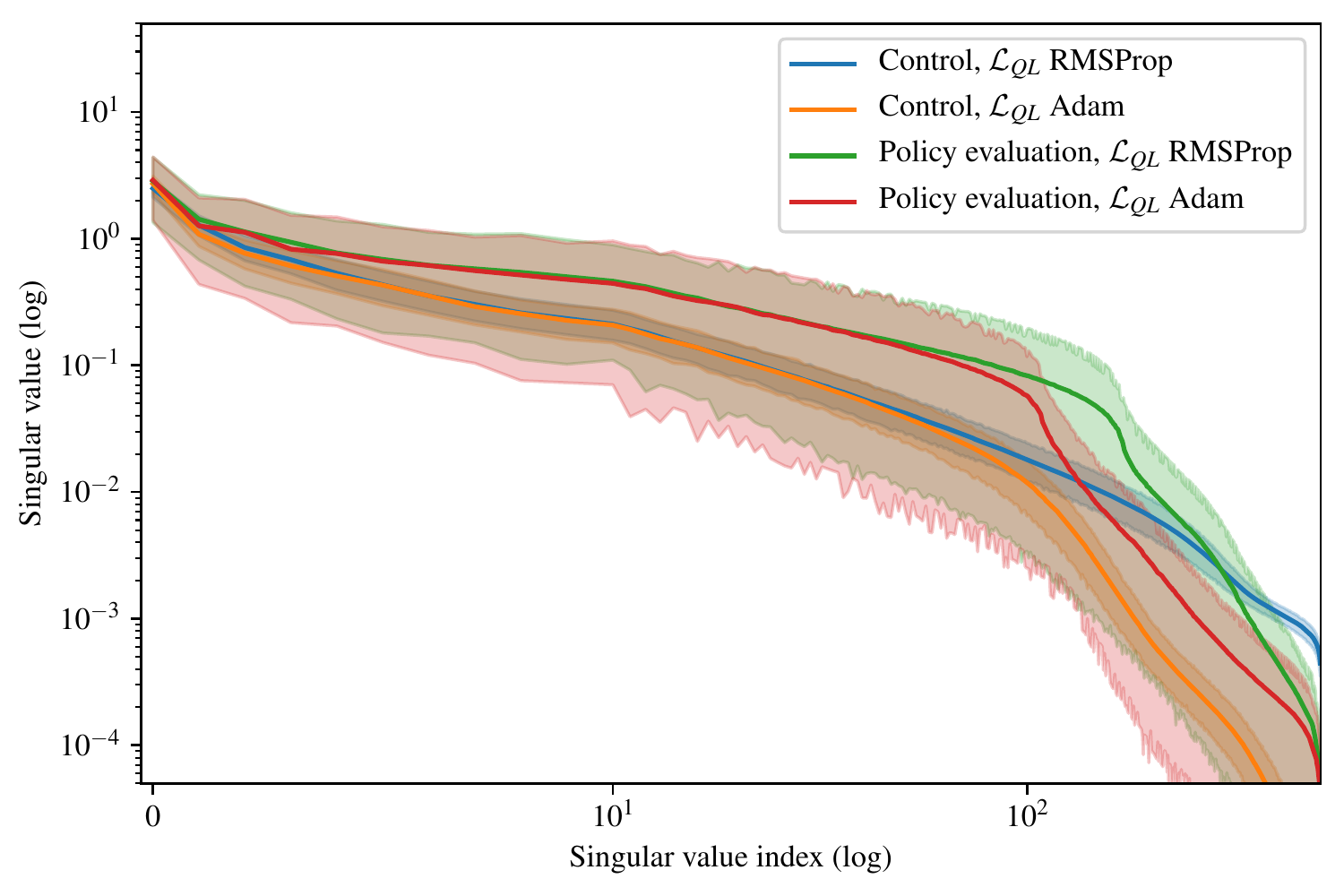}
    \caption{Spread of singular values after 100k iterations. Despite having seen the same amount of data, the control experiments generally have seen fewer unique states, which may explain the observed difference. Shaded regions show bootstrapped 95\% confidence intervals.}
    \label{fig:singular_value_spread}
\end{figure}

Figure \ref{fig:singular_value_spread} shows the spread of singular values after 100k minibatch updates on MsPacman for the Q-Learning objective and Adam/RMSProp. The difference between the control case and policy evaluation supports our hypothesis that policy evaluation initially captures more factors of variation. It remains unclear if the effect of the control case initially having fewer captured factors of variation leads to a form of feature curriculum.

\begin{figure}[h]
    \centering
    \includegraphics[width=0.6\textwidth]{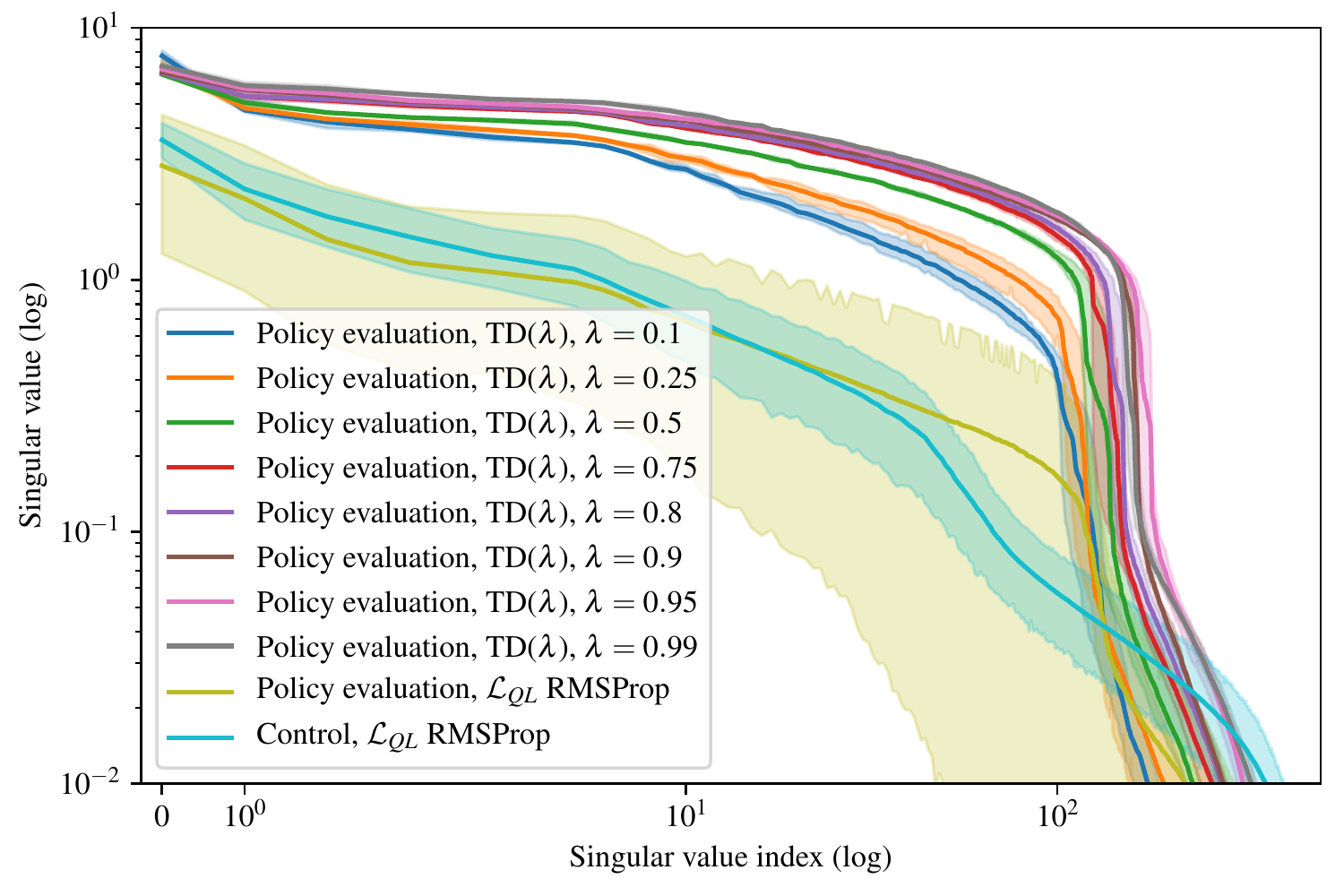}
    \caption{Spread of singular values after 500k iterations. Shaded regions show bootstrapped 95\% confidence intervals.}
    \label{fig:singular_value_spread:tdL}
\end{figure}

Figure \ref{fig:singular_value_spread:tdL} shows the spread of singular values after 500k minibatch updates for TD($\lambda$). Interestingly, larger $\lambda$ values yield larger singular values and a wider distribution. Presumably, TD($\lambda$) having a less biased objective allows the parameters to capture all the factors of variation faster rather than to rely on bootstrapping to gradually learn them.

Note that current literature suggests that having fewer large singular values is a sign of generalization \emph{in classifiers}, see in particular \citet{oymak2019generalization}, as well as \citet{morcos2018importance} and \citet{raghu2017svcca}. It is not clear whether this holds for regression, nor in our case for value functions. Interestingly all runs, even for TD($\lambda$), have a dramatic cutoff in singular values after about the 200th SV, suggesting that there may be in this order of magnitude many underlying factors in MsPacman, and that by changing the objective and the data distribution, a DNN may be able to capture them faster or slower.

\newpage
\subsection{Evolution of TD gain with training}
\label{sec:appendix:figures-of-evolution-of-td-gain}

\begin{figure}[h]
    \centering
    \includegraphics[width=0.7\textwidth]{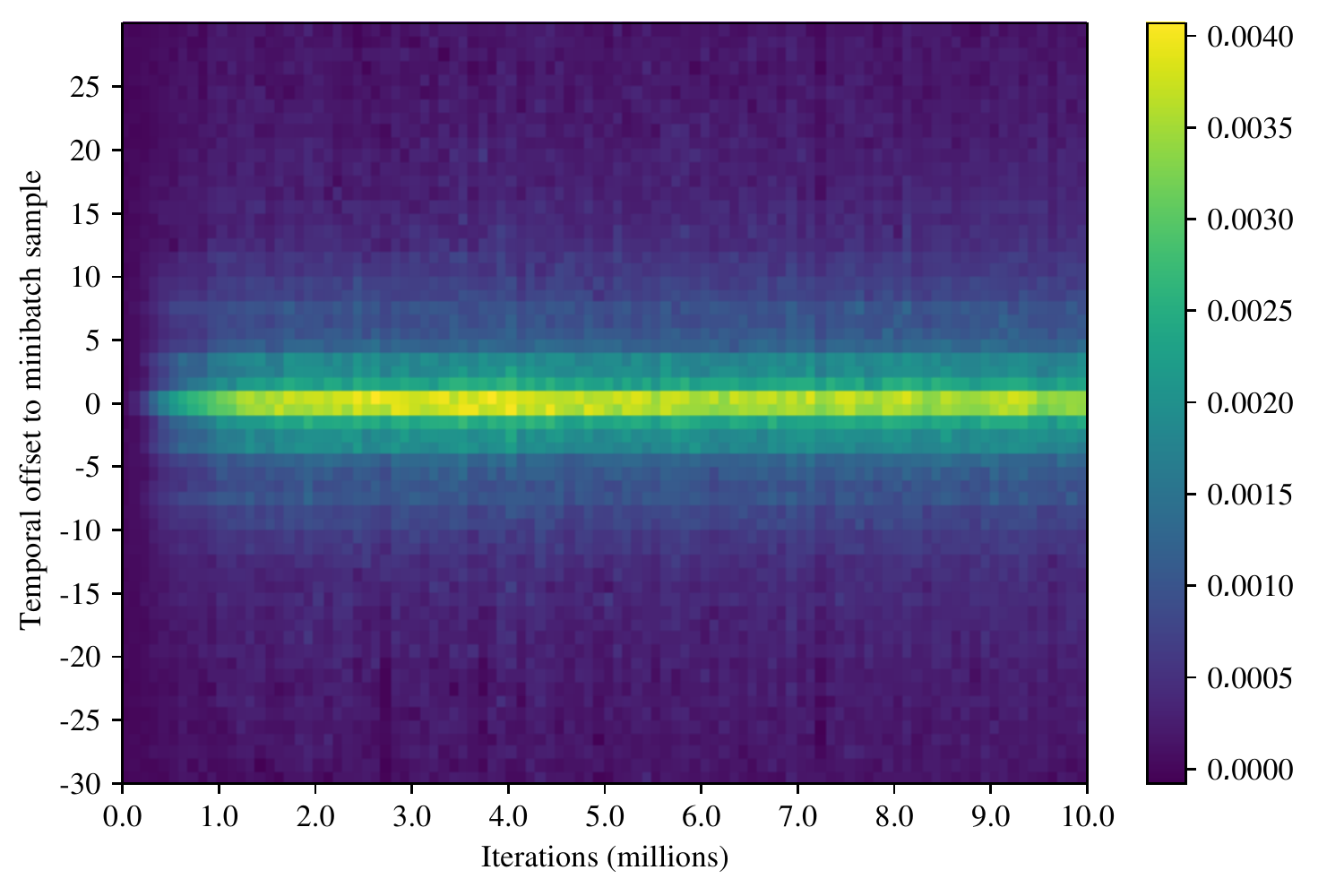}
    \caption{Evolution of TD pointwise loss difference, during training. Control experiment with Adam, MsPacman, averaged over 10 runs. Note that index 0 is excluded as its magnitude would be too large and dim all other values.}
    \label{fig:near_td_gain_evol}
\end{figure}

Figure \ref{fig:near_td_gain_evol} shows the evolution of TD pointwise loss difference during training; in relation to previous figures like Figure \ref{ql_and_sarsa_td_gain}, the $y$ axis is now Fig. \ref{ql_and_sarsa_td_gain}'s $x$ axis -- the temporal offset to the update sample in the replay buffer, the $y$ axis is now training time, and the color is now Fig. \ref{ql_and_sarsa_td_gain}'s $y$ axis -- the magnitude of the TD gain.

\end{document}